\documentclass{article}

\usepackage{times}
\usepackage{graphicx}
\usepackage{amsmath}
\usepackage{amssymb}
 
\usepackage[pagebackref=true,breaklinks=true,letterpaper=true,colorlinks,bookmarks=false]{hyperref}

\usepackage[utf8]{inputenc} % allow utf-8 input
\usepackage[T1]{fontenc}    % use 8-bit T1 fonts
\usepackage{hyperref}       % hyperlinks
\usepackage{url}            % simple URL typesetting
\usepackage{booktabs}       % professional-quality tables
\usepackage{amsfonts}       % blackboard math symbols
\usepackage{nicefrac}       % compact symbols for 1/2, etc.
\usepackage{microtype}      % microtypography
\usepackage{multirow}
\usepackage{algorithm}
\usepackage{algorithmic}
\usepackage{xcolor}
\usepackage{cleveref}
\usepackage{pifont} 
\usepackage{booktabs,subcaption,amsfonts,dcolumn}
\usepackage{caption}

\usepackage{multirow}
\newsavebox{\measurebox}
\usepackage{minibox}
\usepackage{graphicx}

\usepackage{balance}
\usepackage{multirow}
\usepackage{minibox}
\usepackage{array,multirow,booktabs}
 
\usepackage{siunitx}
\usepackage{amsfonts,amssymb}

\usepackage[margin=1.2in]{geometry}
 
\usepackage{xcolor}
\usepackage{xspace}
 \usepackage[switch, modulo]{lineno}
\iffalse
\providecommand{\remarque}[1]{\textcolor[rgb]{1,0,0}{\textit{Rmq: #1}}}
\providecommand{\reponseremarque}[1]{\textcolor[rgb]{0.8,0.2,0}{\textit{Rep: #1}}}

\providecommand{\supp}[1]{\textcolor[rgb]{1,0,1}{(\sout{#1})}}
\providecommand{\comm}[2]{\textcolor[rgb]{1,0,0}{(}\textcolor[rgb]{0.5,0.5,0}{\uwave{#1}}\textcolor[rgb]{1,0,0}{\textit{ #2})}}

\linenumbers
\else
\providecommand{\remarque}[1]{}
\providecommand{\reponseremarque}[1]{}

\providecommand{\supp}[1]{}
\providecommand{\comm}[2]{}

\fi

\usepackage{authblk}

\begin{document}

%%%%%%%%% TITLE
%\title{Improve the Interpretability of Attention: \\ A Fast, Accurate, and Interpretable High-Resolution Attention Model}
\title{BR-NPA: \\ A Non-Parametric High-Resolution Attention Model to improve the Interpretability of Attention}

\author[1]{T. Gomez}
\author[1]{S. Ling}
\author[2]{T. Fréour}
\author[1]{H. Mouchère}

\affil[1]{Nantes University, Centrale Nantes, CNRS, LS2N, F-44000 Nantes, France}
\affil[2]{University of Nantes, Nantes University Hospital, Inserm, CNRS, SFR Santé, Inserm UMS 016, CNRS UMS 3556, F-44000 Nantes, France}

\maketitle

%%%%%%%%% ABSTRACT
\begin{abstract}
The prevalence of employing attention mechanisms has brought along concerns about the interpretability of attention distributions. Although it provides insights into how a model is operating, utilizing attention as the explanation of model predictions is still highly dubious. The community is still seeking more interpretable strategies for better identifying local active regions that contribute the most to the final decision. To improve the interpretability of existing attention models, we propose a novel Bilinear Representative Non-Parametric Attention (BR-NPA) strategy that captures the task-relevant human-interpretable information. The target model is first distilled to have higher-resolution intermediate feature maps. From which, representative features are then grouped based on local pairwise feature similarity, to produce finer-grained, more precise attention maps highlighting task-relevant parts of the input. The obtained attention maps are ranked according to the activity level of the compound feature, which provides information regarding the important level of the highlighted regions. The proposed model can be easily adapted in a wide variety of modern deep models, where classification is involved. %It is also more accurate, faster, and with a smaller memory footprint than usual neural attention modules. 
Extensive quantitative and qualitative experiments showcase more comprehensive and accurate visual explanations compared to state-of-the-art attention models and visualization methods across multiple tasks including fine-grained image classification, few-shot classification, and person re-identification, without compromising the classification accuracy.
The proposed visualization model sheds imperative light on how neural networks `pay their attention' differently in different tasks. 
  % Attention maps are used in many different image processing applications and with various implementations.  Most of the time, its usage aims to improve the performance of the systems.
 % \modif{In this work, we first show that increasing feature map resolution during training and evaluation by reducing stride improves interpretability as the areas on which the model focuses are more located on the object of interest.
 % Combining this with distillation methods yields accurate and interpretable models.}
% In this work, we first propose to distill a low-resolution model into a high-resolution one to improve interpretability as the areas on which the model focuses are more located on the object of interest while maintaining accuracy.
 % Secondly, we propose a model using an algorithm to group features into vectors we call representative vectors to replaces the usual multi-attention neural module. 
\end{abstract}

%%%%%%%%% BODY TEXT
\section{Introduction}
\begin{figure}[!htbp]
\centering
\includegraphics[width=0.7\textwidth]{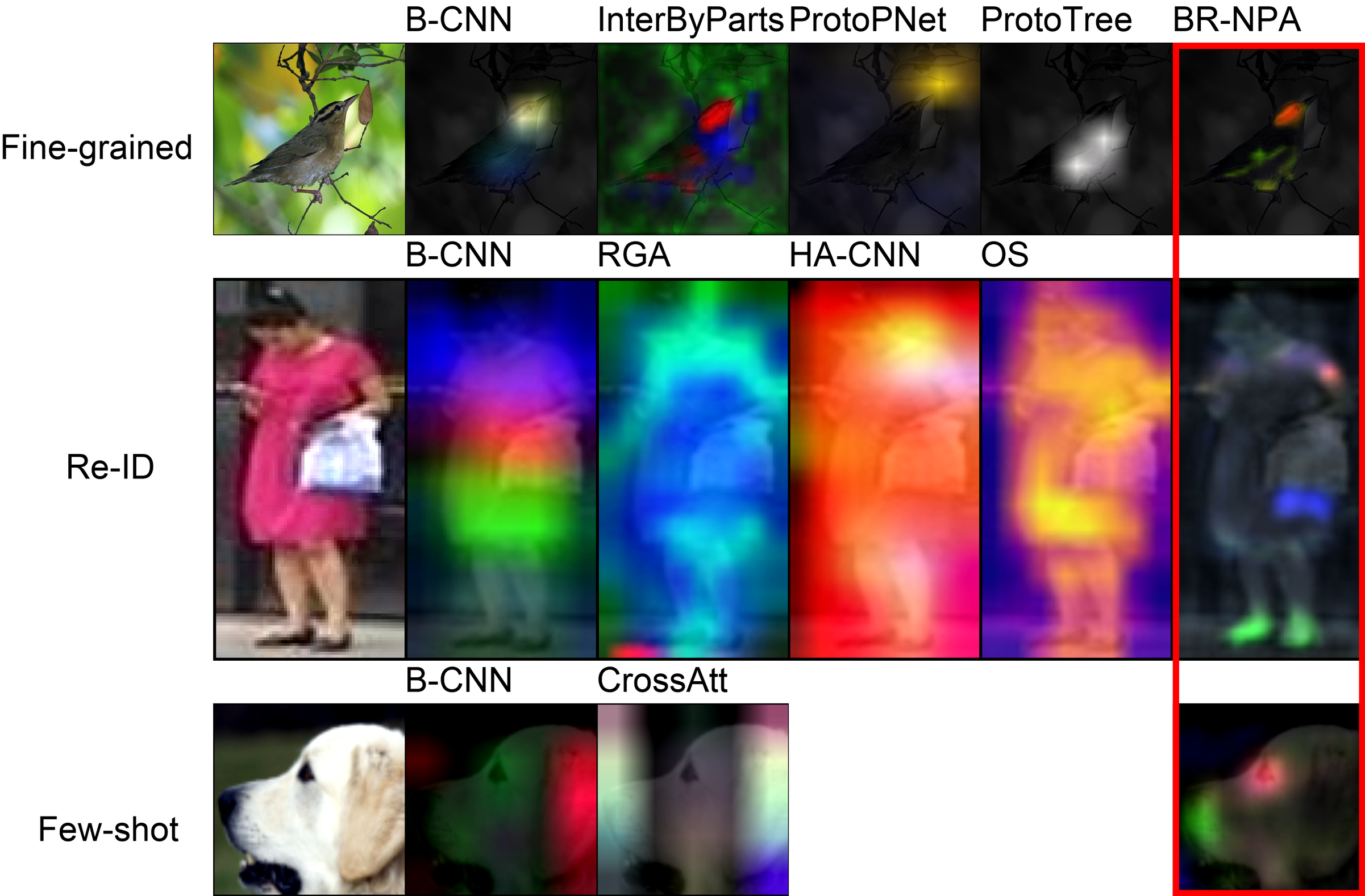}
\caption{
%From left to right: original image, Guided Grad-CAM++, Grad-CAM, Grad-CAM++,  activation map, B-CNN, BR-NPA. 
BR-NPA yields precise attention maps focused on small details and highlights a potential hierarchy among the object's parts relevance. } 
\label{introFig}
\end{figure}

Thanks to the unprecedented booming of deep learning, many applications enjoyed a big leap in performance in the last decade. Nevertheless, these deep learning-based methods are commonly considered `black box' approaches, where the internal functioning is unclear~\cite{gradcam}. As one of the solutions to remedy this lack of interpretability, attention mechanisms~\cite{bahdanau2014neural} are frequently employed for explaining the deep model. Yet to which extent those attention models help in interpreting how the model is operating has recently become a novel debate~\cite{bastings2020elephant,mohankumar2020towards}. An ideal interpretable attention model should not only be able to quantify the important level of a certain part of the neural network but also be capable of identifying precise local regions of the input to provide proper visual guidance regarding which key component is playing a greater role in the decision-making of the corresponding task~\cite{bastings2020elephant}. To this end, in this study, we propose a novel framework that generates high-resolution interpretable attention maps without sacrificing the performance of the target model. More specifically, this framework improves  the interpretability of the usual multi-part attention system designed for deep Convolutional Neural Network models (CNNs) in two ways: (1) we propose to increase the resolution of the output feature maps of a network, and the attention map for finner details ; (2) we propose to replace the common learned attention layer~\cite{bahdanau2014neural} with a non-parametric layer that groups pixels based on their \textit{activity level}, and similarity with others. 
This layer does not require training, which constitutes a simpler function than the standard attention module made of convolutional layers. 
It produces comprehensive and accurate attention maps focusing on semantic parts of the input while sacrificing little to no classification accuracy.

Examples of multi-part attention maps obtained using our method are shown in Fig.~\ref{introFig}.
Our algorithm is designed such that the importance level (in terms of contributions to the final decision-making of the task) of different salient parts of objects could be also friendly visualized via different colors. The $1_{st}$, $2_{nd}$, and $3_{rd}$ attention maps are respectively represented in red, green, and blue. 
Similarly, attention maps generated by state-of-the-art (SOTA) methods are also visualized for comparison. If the method/model produces a single map, it is visualized using a heatmap, \textsl{e.g.,} the OS \cite{OS} method for person re-identification.
%Inspection of the maps shows, for example, that the head of the bird is generally more important than its body in the task of fine-grained classification. 
%Our method can be applied to various problems and produces attention maps adapted to the task at hand. 
%On fine-grained classification and person re-identification, 
\Cref{introFig} shows that contrarily to SOTA  models, BR-NPA precisely highlights small interpretable details consistently across all the studied tasks.
%Experiments on semantic segmentation show similar results to few-shot classification can also be found in supplemental materials. 
 
In the spirit of calls for more interpretable attention models, the focal point of this contribution is to develop a more explainable and interpretable attention model that could be easily plugged into different applications\footnote{Source code is available here \texttt{https://github.com/TristanGomez44/br\_npa}.}. 
Although the proposed attention framework can help to boost performance, we do not focus on contributing to improving the SOTA in standard computer vision tasks, which is beyond the scope of this study.
\comm{Rajouter plan et mentionner les trois tâche et dire que cest des tpaches qui demandent de l'attention etc}{We first discuss related works on three tasks, namely fine-grained classification, person re-identification and few-shot learning.
Being variants of image classification, these tasks are particularly relevant to developing interpretable models with spatial attention.
Next, we describe BR-NPA, followed by extensive experiments before concluding.}

\section{Related works}

We now describe recent works on attention and interpretable architectures.
Generic explanation methods are also mentioned as these also generate interpretable saliency maps and can be compared to explanation maps produced by attention models.

\paragraph{Attention-based architecture for specific tasks:} 

In this work we use fine-grained classification benchmarks extensively as they feature high-resolution images with multi-parts objects, making them adapted to study attention architectures.
Fine-grained classification is an active research area in which various kinds of improvement are proposed.
For example, Chang et al. \cite{mcloss} and Du et al. \cite{multiGran} propose to encourage the CNN to respectively use different feature maps for each class and learn consistent multi-granularity features to improve test accuracy.
This task has also been studied under other frameworks like multi-modality learning \cite{visionAndLanguage} or domain adaptation \cite{inthewild}.
However, our aim is not to improve model performance on fine-grained tasks but rather to improve the interpretability of attention models.
%For example, Chang et al. \cite{mcloss} proposes a custom loss term to encourage the CNN to use different feature maps for each class and the authors observe that, after training, some feature maps roughly correspond to object parts.
%However, this work does not aim at improving the interpretability of attention maps but rather model accuracy.
%Similarly, Du et al. \cite{multiGran} proposes to learn consistent multi-granularity features to improve test accuracy.
%Given the significant amount of work on interpretable fine-grained classification, we rather included interpretable attention architectures instead of other models in our comparison on fine-grained tasks.
For this reason, we focus on existing works about interpretable fine-grained classification and use these in our comparisons.
Many existing attention-based models were developed for fine-grained classification: Lin \textsl{et al.} proposed a bilinear CNN (B-CNN) that employs two distinct CNNs to obtain the feature maps~from the intermediate convolutional layer just before the pooling layer, and further multiply them with outer product~\cite{B-CNN} to explore the coefficient between network features. Inspired by this work, in~\cite{seeBetterBeforeLooking}, it was proposed to generate the attention maps by integrating a convolutional module to the intermediate output feature.
Several similar soft attention layer-based models were also developed for the same purpose, including the Prototypical Part Network (ProtoPNet) \cite{ProtoPNet} that matches learned feature vectors against the feature volume in the manner of a convolution. With a similar recipe, another follow-up work\textsl{, i.e.,} the ProtoTree \cite{prototree} uses a tree of prototypes to make a decision.
By fully exploiting a simple prior of the occurrence of object parts, Huang et al. also showed it was possible to learn to separate an object into parts \cite{interByParts}.
%the multi-attention CNN (MA-CNN) \cite{MA-CNN} that uses a loss term to train an attention layer to group channels according to their similarity, and 

Several methods are proposed to generate part-level crops using reinforcement learning like Fully Convolutional Attention Networks (FCAN) \cite{FCAN} or weakly-supervised learning like Spatial Transformer Networks (ST-CNN) \cite{ST-CNN}, Multi-Granularity CNN (MG-CNN) \cite{MG-CNN}, or MA-CNN \cite{MA-CNN}. 

Also, various neural attention layers were designed for the task of person re-identification \cite{OS,relationAwareReID,harmAttention} or few-shot learning \cite{crossAttFS}.
However, to the best of our knowledge, BR-NPA is the first interpretable model proposed for these tasks, which is why we focus on re-identification and few-shot attention models not designed for interpretability.
%learn3DAtt,dualAtt}. 
On the task of re-identification, Zhang et al. proposed the Relation-aware Global Attention (RGA) module \cite{relationAwareReID} to capture the global structural information. 
In \cite{harmAttention}, the authors suggest to jointly learn the soft and hard attentions along with simultaneous optimization of feature representations.
Omni-Scale (OS) feature learning \cite{OS} was proposed to detect features at multiple scales and dynamically fuse them through a novel aggregation gate.
%Some works have also been introduced to make full use of the particular modalities of the task of person re-identification by exploiting both spatial and temporal clues \cite{learn3DAtt,dualAtt}. 
There are also attention mechanisms that are tailored to few-shot/meta-learning algorithms.
For example, Hou et al. propose a model to better localize relevant object regions and improve sample efficiency \cite{crossAttFS}. 
%Gao_Han_Liu_Sun_2019
%Analogously, Zhu \textsl{et al.} propose to learn diverse and informative parts \cite{multiAttMetaFS}. 
%Recently, a unique module is presented to find relevant frames in a video \cite{ActionRecoFS}. 
Notwithstanding all the appalling attention existing task-specific attention models, one common trait of these works is that they propose attention methods that are tailored for only one typical task. 
On the contrary, we demonstrate that BR-NPA is suited for a variety of tasks.
Moreover, these methods all involve trainable parameters in the attention module whereas BR-NPA's attention maps are generated with a non-parametric algorithm.
Finally, although some of the previous works aimed at designing interpretable models (like \cite{ProtoPNet,prototree,interByParts}), none of them proposed to improve the interpretability of the attention maps, which is the main focus of this work.

\paragraph{Generic explanation methods:} One of the most straightforward ways to identify which local component of the input contributes the most to the model decision-making is the Activation Map (AM) which simply outputs a heatmap showing the average activation across channels. However, not all the activated areas offer necessarily cues for accomplishing the target task, \textsl{e.g.,} classifying a certain category. To overcome this drawback, more interpretable visualization approaches including Class-Activation Map (CAM) \cite{cam} and Grad-CAM \cite{gradcam} were proposed to locate the local task-relevant areas. 
Other methods like Grad-CAM++ \cite{gradcampp} or Score-CAM \cite{scoreCAM} were proposed to improve the interpretability of Grad-CAM.   
Another line of work called RISE \cite{rise} opted instead for a non-gradient-based approach using random masks to find the areas that contribute the most to the prediction.
In~\cite{guidedbackprop}, Guided-Backpropagation (GB) was proposed to visualize the gradients of the input image pixels relative to the predicted output and generate  high-resolution saliency maps that highlight important pixels for the prediction.
This was later improved with noise tunnel-based methods like Vargrad~\cite{varGrad} or Smoothgrad~\cite{smoothGrad}.
%Based on these previous studies, it has also been proposed to combine Grad-CAM with GB to obtain a high-resolution map, \textsl{i.e.,} Guided Grad-CAM, that emphasizes the areas that are related to the prediction~\cite{gradcam}. 

\section{The Proposed Model}
\label{method}
The proposed attention model is composed of (1) extracting high-resolution feature maps; (2) generating representative feature vectors obtained via grouping similar vectors together, and (3) concatenating the representative feature vectors and forwarding them to the classification layer. In this study, the intermediate output of a convolutional network (layer), \textsl{i.e.,}  a tensor with a size of $C\times H\times W$, is defined as the \textit{feature volume} or \textit{feature maps}. Once a spatial position $(h,w)$ is chosen, where $h \in [1, H] , w \in [1, W]$, a vector of dimension $C$ can be extracted, namely, the \textit{feature vector}.

\subsection{High-Resolution Feature Maps \label{reducedStrideSec}}
Given a standard CNN, commonly, the size of the last feature map is significantly smaller than the one of the input. For example, after  feeding an image of $448\times448$ to a ResNet-50 backbone~\cite{resnet} network, the size of the output feature map is reduced to  $14\times14$. This makes it difficult to interpret how the neural network was activated with respect to the image, as each feature vector covers a large and potentially heterogeneous area of the image, \textsl{e.g.} both the object and the background. When visualizing (a part of) the network with features maps of higher resolution, \textsl{i.e.}, with a smaller receptive field, each point within the feature map corresponds to a smaller area of the original that is more likely to be homogeneous.  
To increase the attention map resolution, bi-linear or bi-cubic interpolations are commonly used for visualization. 
Nonetheless, these operations lead to blurry attention maps without details~\cite{gradcam}.
%object_part,pairwise,sun2019finegrained}.
%Another alternative is to consider larger input images as done in \cite{weakSupCompPart}. 
Another alternative is to consider larger input images.
Nevertheless, this method inevitably increases the computation cost and also introduces relevant distortions into original input images during the up-scaling procedure. 
Instead, we proposed to re-adapt the architecture of the CNN backbone used to prevent the reduction of the spatial resolution of feature maps. 
Concretely, this is achieved via reducing the stride of the last downsampling layers in the backbone network. 
For example, on fine-grained classification, we set the strides of the downsampling layers 3 and 4 to 1 (instead of 2), which multiplies the last feature maps resolution by four.
By doing so, the resolution of the feature maps increased from $14\times14$ to $56\times56$ (cf. \cref{details}).
The resolution could be increased even more by also setting the stride of the downsampling layer 2 to 1 too, leading to $112\times112$ attention maps. 
However, in our experiments, we observed that this makes training difficult due to the absence of down-sampling in late layers.
This limits the computation cost increase as only the feature maps of the last layers become larger.
 
%To obtain a higher-resolution model without significant loss of performance, 

In all our experiments, we employ pretrained backbones and this modification implies that the kernels of the last layers will be applied at a scale for which they were not pretrained for. 
As a result, during training, the backbone has to learn features that are adapted to the task but also the new scale, leading to longer training time and longer hyper-parameter optimization, which reduces the efficiency of the transfer learning procedure \footnote{Note that we also experimented with the HR-Net backbone \cite{HRNet} which provides $56\times56$ features maps, but obtained poor performance, probably due to the small number of feature maps provided by HR-Net (16 to 64 depending on the number of layers, whereas ResNet \cite{resnet} provides 512 to 2048 feature maps).}.
%The original HR-Net model achieves good performance by also providing 3 other sets of lower resolution feature maps to the classifier head, amounting to a total of 240 to 960 maps.
%However, using these other sets of feature maps make visualizations more complex as there is 3 more sets of feature maps to visualize.
%The interpretability is also reduced by the existence of the aggregation layer that processes the four sets of feature maps before the final layer, as it may ignore or focus on one set of maps .

To mitigate this issue, the distillation method proposed in~\cite{distillation} is employed. More specifically, a student network with increased resolution is trained to imitate the lower-resolution teacher network. When training the student network, the usual cross-entropy loss was employed jointly with the KL-divergence between the predictions of the student and teacher network to bridge the gap between the two models:
\begin{equation}
    L = \frac{1}{N} \sum\limits_i \alpha\mathrm{CE}(\Tilde{y}_{s},y) + (1-\alpha)\mathrm{KL}(\Tilde{y}_{t} ||\Tilde{y}_{s}) ,
\end{equation}  

where $\Tilde{y}_{t}$, $\Tilde{y}_{s}$ denote the outputs of the teacher and student models, $y$ is the ground truth and $\alpha$ is a parameter that balances the cross-entropy term and the KL divergence term.

\subsection{Non-Parametric Attention Layer \label{diffEmbDiffParts}}

In most cases, the object of interest is composed of several parts, where each part potentially plays a different role in classification. Modern classification models process input and predict corresponding labels in an end-to-end manner, the extraction of important local parts, the modeling of their pairwise correlations, and their contribution to final classification are learned throughout the network. It is thus intuitively appealing to explore the pairwise correlations between the local object parts intermediate feature maps within the network and visualize the part-feature interactions as `attention'. In~\cite{B-CNN}, features corresponding to each object part were aggregated separately to construct a feature matrix, \textsl{i.e.,} a list of feature vectors. The separation of local object parts could be achieved via attention layers composed of multiple convolutional blocks~\cite{seeBetterBeforeLooking}, which is parametric.  
 
With the burgeoning improvement of neural network architectures, SOTA models achieve impressive performances on multiple large-scale benchmarks. Therefore,  the backbone networks utilized in those models should be equipped with the capabilities of separating the local parts. Grounded by well-performing network architecture, instead of incorporating the usual attention neural module, a list of refined feature vectors is formed sequentially by  grouping intermediate feature vectors depending on their \textsl{activation} and their similarity with each other. 
 
In this study, the \textsl{activation} level $a$ of a feature vector $f$ is quantified by its euclidean norm, \textsl{i.e.,} $||f||^2$. The feature vector $f$, whose norm is one of the highest in the corresponding feature volume, is considered to be \textit{active}. Furthermore, a feature vector is considered singular if the other selected features are not similar to it.

 \begin{figure}[!htbp] 
    \centering
    \includegraphics[width=0.8\textwidth]{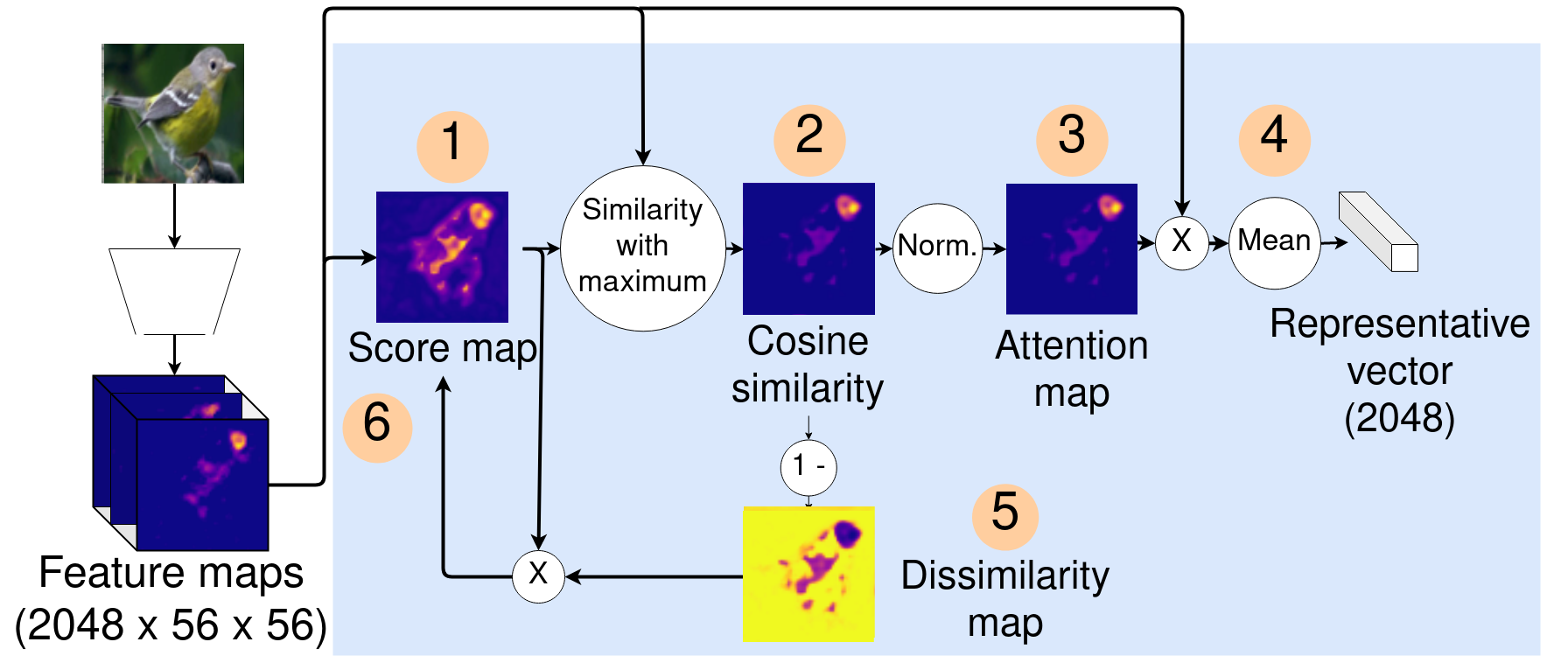}
    \caption{Illustration of the method used to group features without any dedicated module. }
    \label{clustFig} %
\end{figure}  

\begin{algorithm}[tbh]
\caption{Identification of representative vectors $\{\hat{f}_k\}$}
\label{clust}
%\hspace*{\algorithmicindent} 
\textbf{Input : feature vectors $\{f_i\}$}
\begin{algorithmic}
\STATE (1) $a_i \leftarrow ||f_i||^2$
\FOR{$k = 1$ to $N$}
    \STATE $i_{max} \leftarrow \underset{i}{\textrm{argmax}}~a_i$
    \FORALL{$i$}
	    \STATE (2) $s_i \leftarrow \cos(f_i;f_{i_{max}})$
	    \STATE (3) $w_i \leftarrow s_i / \sum_{i'} s_{i'}$
	    \STATE (4) $\hat{f}_{k} \leftarrow \sum\limits_i w_i \times f_i$
	    \STATE (5,6) $a_i \leftarrow (1-w_i) \times a_i$
	   % \STATE
	\ENDFOR
\ENDFOR
\STATE \textbf{return} $\hat{f}_{1},\hat{f}_{2},...,\hat{f}_{N}$
\end{algorithmic}
\end{algorithm}

Alg.~\ref{clust} and Fig.~\ref{clustFig} summarize the proposed feature grouping algorithm. In a nutshell, it selects the top $N$ active singular (\textsl{i.e.,} not similar to each other) feature vectors from the feature volume and further refines them by aggregating them with similar vectors. This algorithm outputs an $N\times M$ feature matrix, where $M$ is the dimension of the vectors. For simplicity, the feature vectors are indexed with only $i \in [0, H\cdot W-1]$, where $H\times W$ is the resolution of the feature map. The $i_{th}$ vector is denoted as $f_i$. The selection of $N$ is discussed in the supplemental materials. 
Experimentally, we observed that using $N>3$ does not improve accuracy significantly and makes visualization more complex.
 
During the first interaction, each $f_i$ is assigned with an activation score $a_i$ equal to its euclidean norm (corresponds to Step $1$ in Alg.~\ref{clust} and Fig.~\ref{clustFig}). Afterward, the vector with the highest norm is selected (Step 2). 
The motivation behind this procedure is that among all the feature vectors in the feature volume of a regular CNN, the ones with higher norms tend to have greater impacts on the final feature vector obtained after the pooling layer \textsl{e.g.,} the average/max pooling. 
When employing a well-performing network architecture for a certain task, the high norm features 
are expected to be more discriminative than the ones with a lower norm. Higher norm features are also supposed to contribute more to the final prediction. Hence, by selecting the high norm vector, we aim at reproducing this phenomenon in our architecture. This intuition is empirically verified in \cref{ablaSect}, where experimental results show that accuracy decreases with random vector selection or with only the $2_{nd}$, $3_{rd}$ ranked features.

After obtaining the feature vector with the greatest norm value $f_{i_{\textrm{max}}}$, it is then refined with other features that are similar to it (Step 2 to 4). Concretely, the cosine similarity $s_i$ between each feature $f_i$ and $f_{i_{\textrm{max}}}$ is computed (Step 2). Subsequently, $\{s_i\}$ is normalized to yield weights $\{w_i\}$ such that $\sum_i w_i = 1$ (Step 3). The first representative feature vector $\hat{f}_1$ is then obtained by computing the weighted average across all the vectors (Step 4). 
This allows stabilizing the representative vector by capturing all the information from similar vectors.
Afterward, a dissimilarity map $1-w_i$ is computed by reversing the weights (Step 5) and the activation scores $\{a_i\}$ are updated by multiplying them by $1-w_i$ for all $i$ (Step 6).
By doing so, it could be ensured that the new chosen location $i_{\textrm{max}}$ and its estimation $f_{i_{\textrm{max}}}$ is different from the ones chosen in the previous iterations,  while still extracting high norm vectors.
%Computing the new $a_i$ by a multiplication of the previous $a_i$ and the dissimilarity $1-w_i$ serves two purposes.
%First, the $a_i$ term should ensure that the next selected feature vector is salient (i.e. has a high norm) and secondly, the $1-w_i$ term ensures that the next chosen vector is dissimilar from the previous ones.

The aforementioned procedure (line 3-8) is repeated for the $k+1_{th}$ feature vectors until $N$ representative  feature vectors are computed. 
Lastly, the obtained $\hat{f}_{1},\hat{f}_{2},...,\hat{f}_{N}$ are concatenated and feed to the final classification layer. 

The weights $\{w_i\}$ corresponding to the top-$N$ representative features serve as attention maps for highlighting local parts that contribute the most to the final decision-making of the task, with the rank of the features indicating the importance level of the respective object parts. It is worth noting that, as there are no extra attention neural layers integrated, our attention model is non-parametric. This scheme can be easily plugged into any modern deep neural network framework, where classification is associated.

\subsection{Discussion}

After step 4, we could repeat the procedure by simply selecting the vector with the second-highest norm to construct a new representative vector.
However, the second-highest norm vector is likely to be spatially close to the highest norm vector, i.e. similar to the highest norm vector.
This implies that the second similarity map is likely to be similar to the first one and that the second representative vector will carry little to no more information than the first one.
To prevent this, we compute a dissimilarity map $1-w_i$ by reversing the weights (Step 5) and update the activation scores $\{a_i\}$ by multiplying them by $1-w_i$ for all $i$ (Step 6). 
The information provided by the dissimilarity map helps to select a vector that is dissimilar from the previous one.
Multiplying the activation scores $a_i$ by the dissimilarities helps to take into account both the high norm criterion and the dissimilarity criterion.
As a consequence, dissimilar high norm vectors will be chosen before dissimilar low norm vectors. In other words, for two vectors that have the same $w_i$, the one with the highest norm will be selected.
We show in \cref{ablaSect} that the 2nd and the 3rd representative vectors indeed carry information that is both salient and different from the information of the 1st vector.

The dimension of the concatenated feature may become large when N increases and hamper optimization. 
However, we use small values of N to keep the model interpretable, preventing any optimization issue.
Moreover, we observe in our analysis of the impact of N (\cref{N}) that even with $N=64$ and a feature vector of size 131072, the accuracy is similar to the accuracy obtained using N=3 and a vector of size $2048\times3=6144$.
This could hint toward the fact that increasing the vector size does not hamper optimization. 
Another explanation is that the objects may have on average only 3 discriminative parts, preventing the 4th to the 64th representative vectors to carry meaningful information, making them ignored during optimization, which results in the same performance as with $N=3$.
An inspection of the weights given to the features of these vectors would confirm that hypothesis but we leave this for future work.

\section{Experiments \label{results}}

The proposed framework was first evaluated on various computer vision tasks on relevant SOTA benchmarks, including the fine-grained image classification, \cite{CUB,aircraft,StanfordCars} few-shot classification \cite{tieredimagenet}, and person re-identification \cite{market1501}\footnote{More visualisations are available in supplementary material.}.
Furthermore, extensive ablation studies were conducted to validate the chosen strategies. Lastly, a quantitative evaluation of the attention maps using dedicated metrics was performed to demonstrate the performance of the proposed framework.

%For performance comparison regarding visualization, 
We compared the proposed BR-NPA attention model with SOTA models from the literature like B-CNN \cite{B-CNN}, ProtoPNet \cite{ProtoPNet}, ProtoTree \cite{prototree}, Interpretability by parts (IBP) \cite{interByParts}, RGA \cite{relationAwareReID}, HA-CNN \cite{harmAttention}, OS \cite{OS}, and Cross-Attention (CA) \cite{crossAttFS}, and also SOTA post-hoc algorithms dedicated to model visualization including Grad-CAM~\cite{gradcam}, Grad-CAM++~\cite{gradcampp}, Guided backpropagation~\cite{guidedbackprop}, Score-CAM \cite{scoreCAM}, RISE \cite{rise}, VarGrad \cite{varGrad}, SmoothGrad \cite{smoothGrad}, and also the baseline activation map (AM) method. %These methodologies were selected as most CNN attention maps in the literature were generated using one of them.
The attention models were trained using the author's code and we used Captum \cite{captum} for the generic visualization methods.
For fine-grained classification, we also report for reference the performance of 2-level attn. \cite{twoLevelAtt}, MG-CNN \cite{MG-CNN}, FCAN \cite{FCAN}, ST-CNN \cite{ST-CNN}, and MA-CNN \cite{MA-CNN}. 
These models were not included in the qualitative or quantitative evaluations of the saliency maps because they are hard attention models that produce bounding boxes, preventing comparison with the soft attention maps produced by BR-NPA, but they are still attention models whose performance is interesting to compare to.

%Every post-hoc algorithm was applied to the three tasks, where for each task, a regular CNN was trained, and the algorithms were applied on it.
B-CNN is a simple baseline architecture that makes minimal hypotheses about the task at hand, making it appropriate to be applied to other tasks than the one it was developed for. Therefore, among the aforementioned non-post-hoc models, there is only B-CNN was trained on all considered tasks, since the other attention models were developed/designed for a certain dedicated task.
On the other hand, since the post-hoc algorithms are generic methods, each of them was applied to all three tasks.
The visualized results (maps) of the post-hoc algorithms were obtained using a regular CNN without attention trained with the same setting as the attention models.

%The models which performance is reported from the literature in the fine-grained experiments were not applied to person re-identification or few-shot classification are they are complex methods specifically tailored for fine-grained classification. Only the variant of B-CNN proposed by \cite{seeBetterBeforeLooking} was re-implemented and showed poor performance on re-identification and few-shot classification.
%Also, the visualisations produced by these models come either from AM (MG-CNN) or a neural attention module (B-CNN, FCAN, MA-CNN, ProtoPNet). ST-CNN and 2-level attention produce no saliency map as the first directly predicts the crop to apply and the latter uses an object detector on random image patches to filter out the ones showing background.
%Therefore, visualisations from AM and neural attention module are provided, along with visualisations obtained from generic methods like Grad-CAM, Grad-CAM++, and Guided Grad-CAM++, to obtain a comprehensive and concise representation of the methods in the literature.
 
\subsection{Implementation Details \label{details}}

As mentioned in \cref{diffEmbDiffParts}, we chose $N=3$ for BR-NPA as it is a good trade-off between accuracy and interpretability.
More detail about the selection of $N$ can be found in supplementary materials.
As the number of attention maps generated for an image varies greatly among other models (\textsl{e.g.}, $32$ for IBP, up to $2000$ for ProtoPNet),  we selected $3$ attention maps among the ones produced by each model.
This ensures concise and fair qualitative comparison among models. For visualization, each of the 3 attention maps is mapped to one channel of an RGB image (details are given below).
Only the B-CNN model was trained using $3$ parts, as using a larger number of object parts tends to reduce its performance\footnote{More details can be found in supplementary materials.}. For other models, we kept the same number of part/attention maps as reported in the original papers.
To select the attention maps, we adopted different strategies.
For the ProtoTree and ProtoPNet, we chose the most active maps, \textsl{i.e.}, the maps that have the highest similarity score with the input image, as they are the most relevant regarding the target task.
For RGA and HA-CNN we used the attention maps generated by the last $3$ layers, as they are the closest to the decision-making layers.
Once the $3$ maps are selected, they were visualized by showing the $1_{st}$, $2_{nd}$ and $3_{rd}$ maps respectively in red, green, and blue.
More precisely, the red, green, and blue channels of each pixel are were to represent its contribution \textsl{w.r.t.} the $1_{st}$, $2_{nd}$ and $3_{rd}$ attention maps. The three channels were also multiplied by the norm of the corresponding feature vector, to make salient pixels brighter.
The OS model and the saliency algorithms only generate one attention map, they were visualized using a simple heatmap.

%\remarque{Tristan : Add map selection criteria for CrossAtt and InterByParts.}

%\comm{Tristan : It is strange to detail B-CNN and not the other models and moreover it is not necessary because these details and the ones of all the other models can be found in the papers of these models}{
%All models were trained in the setting provided by the authors except B-CNN.
%To adapt B-CNN~\cite{B-CNN} for visualization, the features extracted from the backbone network were fed into a ResNet basic block \cite{resnet} followed by a $1\times1$ convolution and a ReLU activation which output is used as the attention maps. The basic Resnet block is composed of $2$ sequential convolutional layers (with the batch normalization and ReLU activation), and the residual connection followed by a final ReLU activation.
%Once obtained the attention maps, each map is multiplied with the feature volume before applying the pooling layer. In total $C'$ feature vectors could be obtained (same number as attention maps). Finally, the vectors were concatenated and passed to a softmax layer.}

%BR-NPA designed to be a high resolution model, which is why the stride in the last downsampling layers of the original backbones were reduced to increase feature maps and attention map size (cf. \ref{reducedStrideSec}). 
%More details are provided in the following paragraphs.
%To achieve \modif{more precise visualisation}{better training} with higher-resolution, network distillation was conducted as detailed in \cref{reducedStrideSec}. 

For all experiments and tasks, hyper-parameters were estimated with the Tree-structured Parzen Estimator (TPE) from the Optuna framework \cite{optuna}, except in \cref{ablaSect}, where we used the default hyper-parameter values detailed in supplementary materials. 
Details of how the proposed attention was adjusted and employed for different tasks are given below.
%\supp{, where a similar recipe was applied for B-CNN}.

\textbf{Fine-grained image classification:} For both BR-NPA and the baseline model B-CNN, the ResNet-50 \cite{resnet} pre-trained on the ImageNet dataset \cite{imagenet} was used as the backbone network since it is one of the most commonly used network architectures in various domains.
%\footnote{\comm{Tristan : I don't think this is necessary given related papers only use one backbone too (and its mostly, the one we are using}{Other adapted network architectures were also tested. More details and results are given in the supplemental materials.}}
To obtain higher-resolution feature maps, the stride of the downsampling blocks in the $3_{rd}$ and $4_{th}$ layer of the ResNet-50 backbone network was set to $1$ instead of $2$, which increased the feature maps from $14\times14$ to $56\times56$. The teacher model used for distillation is a BR-NPA with an unmodified backbone, i.e. with a $14\times14$ resolution.

\textbf{Few-shot classification:} The E3BM model proposed in~\cite{e3bm} is one of the SOTA meta/few-shot classification models. It was thus employed as a backbone with the studied attention model, \textsl{i.e.,} B-CNN, CA, and BR-NPA for more comprehensive visualization. To compare BR-NPA with B-CNN and CA, the average pooling layer in the backbone in E3BM was replaced with the studied attention layers.

For B-CNN and BR-NPA, the original  classification layer was also extended to take the concatenation of the 3 representative features as input. 
No such modification was necessary for CA as it does not extend the size of the feature vector produced.
The model was then trained as described in \cite{e3bm} one of the standard few-shot classification benchmark datasets TieredImagenet \cite{tieredimagenet} using pretrained weights provided by authors to reduce training time. 
In line with E3BM, the ResNet-12 was used. The stride of the downsampling blocks in $2_{nd}$ and $3_{rd}$ layer of the ResNet-12 were set to $1$ instead of $2$. As a result, the resolution of the feature maps was enlarged from $5\times5$ to $21\times21$. The teacher model used for distillation is the pretrained E3BM model which weights were provided by the authors.
 
\textbf{Person re-identification:} Similar to the setup for the few-shot classification task, one of the best-performing models proposed in~\cite{DG-Net}, \textsl{i.e.,} DG-Net, was adapted with the studied methods. 
For BR-NPA and B-CNN, the average pooling layer in the identification module used by \cite{DG-Net} was removed and replaced with the attention layer, and the final dense layer was also expanded to match the new feature vector size.
To prevent a size mismatch between the weights for the original dense layer with the feature vector extraction ($512\times2048$) and the expanded dense layer ($512 \times 2048\cdot N $), we duplicated the weights $N$ times along the feature axis. This procedure was used to train DG-Net in combination with both BR-NPA and B-CNN.
For RGA, HA-CNN, and OS, the original backbone was replaced by the CNN proposed by these methods, trained in the same setup as DG-Net.
Then, the model was trained as elucidated in~\cite{DG-Net} on the popular Market-1501 dataset \cite{market1501}.
It has to be emphasized that the DG-Net has a complex architecture that is difficult to train from scratch. Thus, we utilized the pretrained weights provided by the authors and modified the DG-Net architecture to speed up the training procedure. 
%However, the proposed method extracts $N$ feature vectors, whereas the original model extracts only one feature vector for classification.

The ResNet-50 was used as the backbone for the task of person re-identification to be consistent with \cite{DG-Net}.
The strides of the $2_{nd}$ and $3_{rd}$ layers of the ResNet-50 were set to $1$ instead of $2$, knowing that the stride of the $4_{th}$ layer was already set to $1$ by the original authors. This increased the resolution of the feature maps from $16\times8$ to $64\times32$. All models were trained with the same distillation procedure and the same teacher model as the original DG-Net model.

\subsection{Experimental Results}
\textbf{Fine-grained image classification:} The performances of the proposed model and SOTA visualization models in terms of accuracy on three fine-grained datasets, including the CUB-200-2011 \cite{CUB}, FGVC-Aircraft \cite{aircraft}, and the Standford cars \cite{StanfordCars} datasets are depicted in Table~\ref{sota}. It is shown that BR-NPA achieves SOTA performances, and outperforms the other interpretability-oriented methods on the three datasets.  

\begin{table}[!htbp] 
\centering
\begin{tabular}{c|c|c|c|c}
\toprule
Method  & Attention & CUB & FGVC & Stanford cars  \\
\midrule
2-level attn. \cite{twoLevelAtt} & Hard & 77.9 & - &  -\\
%Neural const. \cite{neuralActivationConst} & 81.0 & - & - \\
MG-CNN \cite{MG-CNN}& Hard & 81.7& - & -\\
FCAN \cite{FCAN} & Hard &82.0& - & -\\
ST-CNN \cite{ST-CNN}& Hard &84.1& - & -\\
MA-CNN \cite{MA-CNN}& Hard &85.4 & 88.4 & 91.7 \\ 
\hline
IBP \cite{interByParts} & Soft &81.9 & 87.9 & 89.5 \\
ProtoTree \cite{prototree} & Soft &82.1 & 81.3 & 85.7 \\
B-CNN \cite{B-CNN} & Soft & 84.1 & 84.2 & 91.3\\
B-CNN (our impl.) & Soft &82.9 & 88.6 & 89.9\\
ProtoPNet \cite{ProtoPNet} & Soft &84.8 & 84.0 & 85.8\\
%InterByParts \cite{interByParts} & 84.2 & $\mathbf{92.4}$ & $\mathbf{93.6}$ \\
%InterByParts \cite{interByParts} & 47.8 & 76.8 & 66.7 \\
%FV-CNN \cite{FV-CNN} & - & 88.4& -\\
%RA-CNN \cite{RA-CNN}& 85.3 & 88.2 & $\mathbf{92.5}$ \\
\hline
BR-NPA & Soft & $\mathbf{85.5}$ & $\mathbf{89.6}$ & $\mathbf{92.2}$ \\ %92.2 on cars now
\bottomrule
\end{tabular}
\caption{\label{sota} Performance for task of fine-grained classification.}
\end{table}

\begin{figure*}[tb]
\centering
\includegraphics[width=\textwidth]{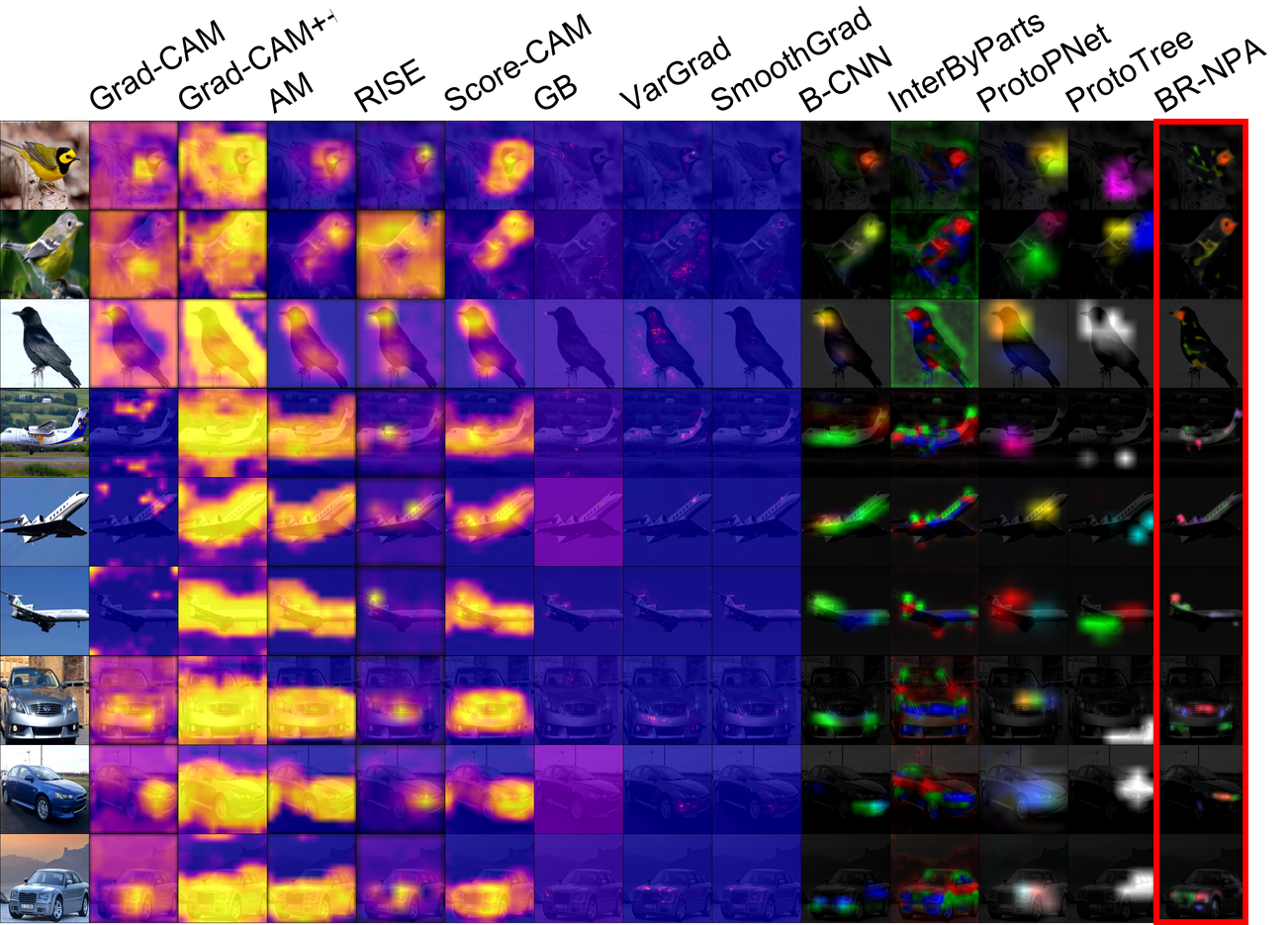}
%\caption{From left to right : original image, Guided Grad-CAM++ \cite{gradcampp}, Grad-CAM \cite{gradcam}, Grad-CAM++ \cite{gradcampp}, AM \cite{WeaklySupervisedLearningOfObjectPart,MA-CNN}, B-CNN \cite{B-CNN,seeBetterBeforeLooking}, BR-NPA. The first four columns indicate saliency with a yellow color, whereas the last two rows indicate saliency using brightness and use color to indicate by which map the pixel was attended. Red, green and blue represent the $1_{st}$, $2_{nd}$ and $3_{rd}$ maps  respectively. } 
\caption{Comparison of different visual explanations with BR-NPA for fine-grained classification on 3 datasets: CUB-200-2011  lines 1-3, FGVC-Aircraft line 4-6, and the Standford cars lines 7-9. BR-NPA yields detailed attention maps focused on semantic parts of the object whereas the other methods either produce blurry maps covering the whole object or extremely sparse maps making it difficult to identify the important areas.
%The Plasma color map from Matplotlib\cite{matplotlib} is used for methods generating one attention map, whereas methods producing several are visualized using brightness for saliency and color to indicate by which map the pixel was attended. Red, green and blue represent the $1_{st}$, $2_{nd}$ and $3_{rd}$ maps respectively. 
%Note that, except for BR-NPA, the attention maps generated by the studied here are unordered.
}
\label{fig:fineGrainedComp}
\end{figure*}

Selected visualizations of the attention maps are presented in Fig.~\ref{fig:fineGrainedComp}. Among all the images, the attention maps produced by BR-NPA highlight more precise details of the objects that capture well the inter-class unique characteristic. For example, in rows 4-6, highlighted finer-grained local regions are most of the time the tails or wings of the airplanes, which help to differ different airplane models. 
Similarly, the other attention models are also able to identify similar local important parts, but they are generally less precise and consistent. 
Conversely, the highlighted regions from Grad-CAM, Grad-CAM++, AM, RISE, and Score-CAM approaches are significantly coarser, which cover the entire objects, while the ones from Guided-Backpropagation, VarGrad, and SmoothGrad are extremely sparse, making it difficult to identify the important areas and also sometimes spread out randomly all over the images. 
None of those attention regions provide precise information that helps the decision-making of fine-grained classification.

%In Table~\ref{sota} we report the accuracy of a high-resolution B-CNN and a high resolution BR-NPA using distillation and compare them with interpretability aimed methods on the three datasets. 
%Our model yields SOTA performance over interpretability-oriented methods on the CUB-200-2011 and the FGVC-Aircraft datasets. 

\begin{figure*}[tb]
\centering
\includegraphics[width=\textwidth]{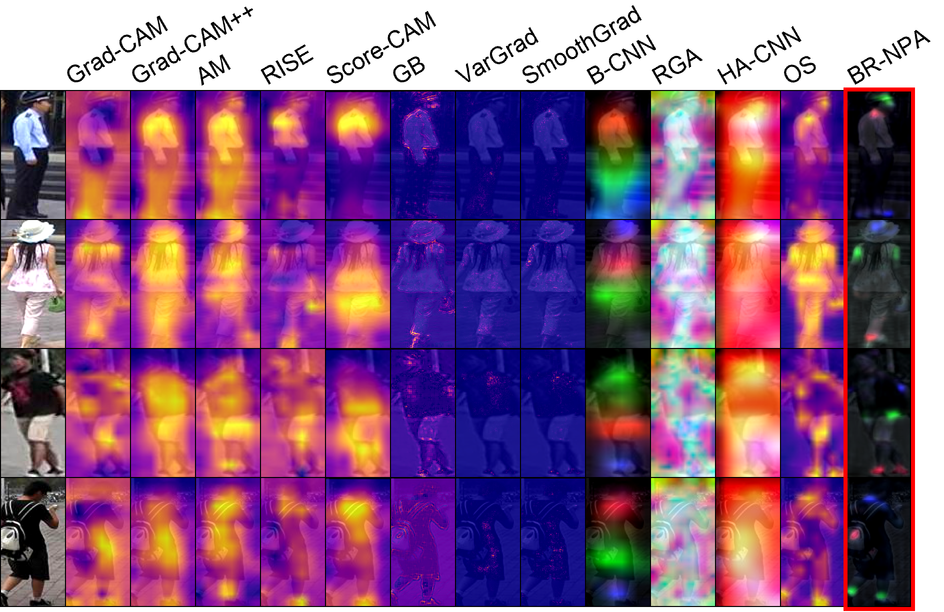}
%\caption{From left to right : original image, Guided Grad-CAM++ \cite{gradcampp}, Grad-CAM \cite{gradcam}, Grad-CAM++ \cite{gradcampp}, AM \cite{WeaklySupervisedLearningOfObjectPart,MA-CNN}, B-CNN \cite{B-CNN,seeBetterBeforeLooking}, BR-NPA. The first four columns indicate saliency with a yellow color, whereas the last two rows indicate saliency using brightness and use color to indicate by which map the pixel was attended. Red, green and blue represent the $1_{st}$, $2_{nd}$ and $3_{rd}$ maps  respectively. } 
%\remarque{Tristan : I don't know if I should write something here, as it would be the same thing as in the previous figure}
\caption{Comparison of different visual explanations with BR-NPA for the re-identification task on the Market-1501 dataset.\label{fig:reidComp}}
\end{figure*}  

\begin{figure*}[tb]
\centering
\includegraphics[width=\textwidth]{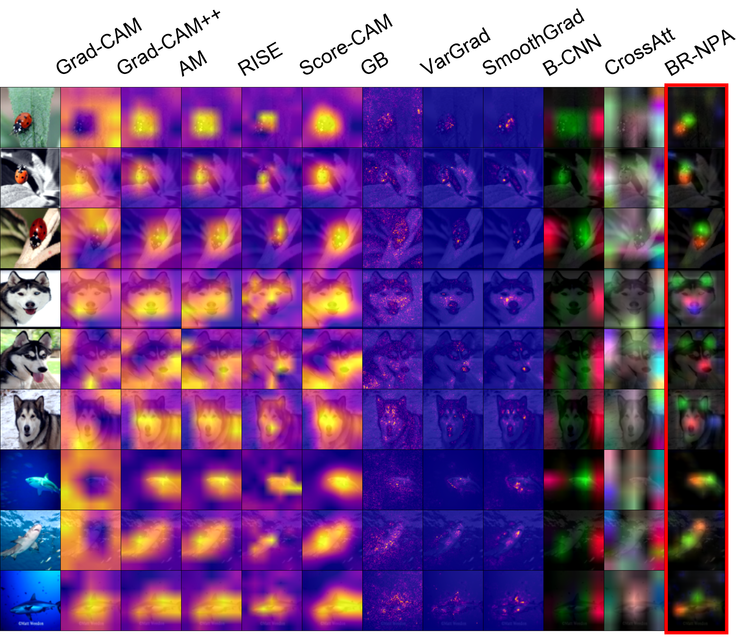}
%\caption{From left to right : original image, Guided Grad-CAM++ \cite{gradcampp}, Grad-CAM \cite{gradcam}, Grad-CAM++ \cite{gradcampp}, AM \cite{WeaklySupervisedLearningOfObjectPart,MA-CNN}, B-CNN \cite{B-CNN,seeBetterBeforeLooking}, BR-NPA. The first four columns indicate saliency with a yellow color, whereas the last two rows indicate saliency using brightness and use color to indicate by which map the pixel was attended. Red, green and blue represent the $1_{st}$, $2_{nd}$ and $3_{rd}$ maps  respectively. } 
\caption{Comparison of different visual explanations with BR-NPA for the few-shot classification task on the TieredImagenet dataset\label{fig:fewshotComp}.}
\end{figure*}  

\textbf{Person Id-reidentification:} The accuracy of DG-Net equipped with different attention methods on one of the most common benchmarks Market-1501~\cite{market1501} is illustrated in Table~\ref{perfReID}. Compared to other attention schemes, the lower-resolution version of BR-NPA achieves the best performance. 
Its performance is slightly poorer compared to the original DG-Net. Yet, as aforementioned, the complex architecture makes it hard to train from scratch, and we hence employed the pre-trained weights. One of the possible reasons for the slight performance drop could be that the pretrained network was not well adapted to our attention framework, and therefore are more prone to a local optimum.
 
Comprehensive attention maps are shown in Fig.~\ref{fig:reidComp}. Similar to fine-grained classification, the attention maps produced by BR-NPA emphasize local important  details of the clothes on the person that are momentous for the ID re-identification regarding the significant intra-class variations across different cameras. For example, highlighted regions include shoes (rows 1-4), a logo on the bag (row 4), pockets (row 3), or epaulet (row 1), which help to differentiate different persons. B-CNN and OS are only able to focus on larger parts of the person, \textsl{e.g.,} the upper body (rows 1-3) or the whole handbag (row 3), and the RGA and HA-CNN methods barely focus on the body of the person. 
%\comm{Tristan : This is a rephrasing of the previous paragraph on fine-grained recognition. Do you think its a problem ?}
Similarly to fine-grained recognition, the highlighted regions indicated by the Grad-CAM, Grad-CAM++, AM, RISE and Score-CAM approaches generally cover the entire objects, while the ones from Guided-Backpropagation, VarGrad, and SmoothGrad are extremely sparse, making it difficult to identify the important areas.
%\supp{The particular noisy aspect of Guided Grad-CAM++ on this task could be due to the high level of blur in the input compared to the images utilized in fine-grained classification. Moreover, the Guided-backpropagation-like methodologies have been proven to focus on edges\cite{sanityCheckForSalMap}, which are less robust to blurry, lower-resolution input.}
%\begin{table}[!htbp]
%    \centering
%    \begin{tabular}{c|c|c|c}
%        \hline 
%        Method &  Attention & Resolution & Accuracy \\
%        \hline 
%        \multirow{4}{*}{DG-Net \cite{DG-Net}}   & \xmark & $16\times8$ & $94.8$ \\
%         & B-CNN & $16\times8$ & $78.4$ \\
%         & BR-NPA & $16\times8$  & $93.6$ \\
%         & BR-NPA & $64\times32$  & $88.2$ \\
%        \hline 
%    \end{tabular} 
%    \caption{Performance for task of person re-identification.\label{perfReID}}
%\end{table}  
\begin{table}[!htbp]
    \centering
    \begin{tabular}{c|c|c|c|c}
        \toprule
        \multirow{2}{*}{Model} & Reference        & \multirow{2}{*}{Attention} & \multirow{2}{*}{Resolution} & \multirow{2}{*}{Accuracy} \\
              & Accuracy    &           &            &     \\
        \midrule
        \multirow{5}{*}{DG-Net } & \multirow{5}{*}{$94.8$} & B-CNN        & $16\times8$ & $80.1$ \\
                                 &                         & HA-CNN       & $10\times4$ & $86.6$ \\
                                 &                         & OS           & $16\times8$ & $91.2$ \\
                                 &                         & RGA          & $16\times8$ & $93.0$ \\
        \cline{3-5}
                                &                          & \multirow{2}{*}{BR-NPA (ours)}& $16\times8$ & $\mathbf{93.6}$\\ 
        &                                                   &                              & $64\times16$ & $88.1$\\ 
        \bottomrule
    \end{tabular}
    \caption{Performance for the task of person re-identification.}
    \label{perfReID}
\end{table} 

Let alone the negligible performance drops compared to the original DG-Net, it is showcased that our BR-NPA could be easily plugged into a complex pre-trained architecture without suffering from a significant loss of accuracy while providing insightful visualization.
 
\textbf{Few-shot classification:} We tested the combination of the E3BM model with 3 attention models: B-CNN, CA, and the BR-NPA modules. Each combination was evaluated in a 5-shot 5-way few-shot setting on the CIFAR-FS dataset~\cite{cifarFS}. Mean accuracies are summarized in Table~\ref{perfFewShot}. 
As observed, by integrating the $5\times5$ resolution version of BR-NPA, the accuracy of the original E3BM model is improved, and both $5\times5$ and $21\times21$ resolution versions of BR-NPA provide performance superior to the version with B-CNN or CA 
%The E3BM model uses a popular method in two steps to train a model for few-shot classification \cite{modelAgnos}. During meta-training, the backbone weights were updated \supp{trained} such that the classification layer weights do not need many learning steps to generalize to a new task from random initialization. 
%During meta-evaluation, the backbone weights are fixed and the classification weights re-trained for each new test task.
%As integrating BR-NPA into the E3BM model only modifies the classification layer, we were able evaluate BR-NPA without necessitating a meta-training phase. With BR-NPA, the accuracy was boosted to $85.6\%$, which demonstrates that  
%BR-NPA is able to obtain performance close to the original model and its performances are superior to the ones with B-CNN and CA
This shows that BR-NPA is a versatile module that can be integrated into an existing pretrained architecture and maintain the model performance.

\begin{table}[!htbp]
    \centering
    \begin{tabular}{c|c|c|c|c}
        \hline 
        \multirow{2}{*}{Model} & Reference               & \multirow{2}{*}{Attention}    & \multirow{2}{*}{Resolution} & \multirow{2}{*}{Accuracy} \\
                               & Accuracy                &                               &                             &  \\
        \hline 
        \multirow{5}{*}{E3BM } & \multirow{5}{*}{$85.8$} & CAM                           & $5\times5$                 & $81.9$ \\
                               &                         & B-CNN                         & $5\times5$                 & $83.2$ \\
        \cline{3-5}
                               &                         & \multirow{2}{*}{BR-NPA (ours)}& $5\times5$                 & $\mathbf{85.9}$\\ 
        &                                                &                               & $21\times21$               & $85.3$\\ 
    \end{tabular}
    \caption{Performance for the task of few-shot classification.}
    \label{perfFewShot}
\end{table} 

Representative visualized maps are depicted in in \cref{fig:fewshotComp}. On one hand, as opposed to fine-grained classification and person re-identification, in the task of few-shot classification, attention maps generated by BR-NPA highlight~slightly larger areas of the object, while most of the local parts are still well determined and are distinguishable from each other. For instance, BR-NPA separates the body from its head (rows 1-3 and 7-9) or the ears from the nose (rows 4-6), which are all key relevant components for recognizing the corresponding categories. These observations demonstrate the feasibility of adapting BR-NPA for few-shot learning classifications. Interestingly, unlike the visualization maps obtained in the task of fine-grained classification, the model tends to focus on the entire objects instead of finer regions in a few-shot regime, which also verifies the interpretability of the proposed models across tasks. On the other hand, B-CNN is not able to differentiate different local parts as it only generates one meaningful attention map (in green), the other one (in red) exhibiting a seemingly degenerate behavior. 
Similarly, CA does not differentiate object parts either and exhibits an even more degenerate behavior.
Again, the highlighted regions from Grad-CAM, Grad-CAM++, AM, RISE and Score-CAM approaches generally cover the entire objects, while the ones from Guided-Backpropagation, VarGrad, and SmoothGrad are extremely sparse, making it difficult to identify the important areas and also sometimes spread out randomly all over the image.

%indicate the ranking of discriminate power of different features in terms of their contributions to the final task decision-making.
\textbf{Comparisons of attention maps obtained from different tasks:} The attention maps obtained from different tasks highlight the target object at different granularity-level. 
For the task of fine-grained classification and person re-identification, our method yields more precise attention maps that focus on finer object parts. 
Differently, for few-shot classification, the proposed model tends to focus on the entire object as the objective of few-shot is to distinguish between categories instead of sub-categories with only a few samples per unseen class. It is demonstrated that our model can generate interpretable visualized maps according to the goal of the task, which could provide important insights for the development of different methodologies, \textsl{i.e.,} different network architectures that pay more attention to a certain type of local areas, regarding the goal of the target task.

\subsection{Ablation study \label{ablaSect}}

%\protect\footnote{Due to limited space, only results for ablation study of fine-grained classification are presented, others are in the supplemental materials.}

\textbf{Impact of active feature selection and refinement strategy:} Three ablative studies were conducted to explore the impacts of the proposed strategies of active feature vectors selection and the feature refinement on the performances, including  
(1) the ablative model where the feature vectors are uniformly and randomly selected among all available vectors instead of utilizing the active selection strategy; 
(2) the ablative model without the refinement phase, which simply concatenates all the active vectors and passes them directly to the classification layer; 
(3) a model with random selection and no refinement phase. During this ablation experiment, the aforementioned 3 ablative models along with the original model were trained with reduced strides to increase resolution from $14\times14$ to $56\times56$ (cf. \cref{details}) without distillation or hyperparameter optimization for the sake of simplicity. 
Results concluded in Table~\ref{ablation} on the CUB-200-2011 dataset verifies the validity of both the active feature selection and refinement strategies.   

\begin{table}[!hptb]  
 \centering
\begin{tabular}{c|c|c}\toprule
Raw vector selection method & Refining & Accuracy   \\ 
\hline
Random & No & $0.5$ \\
Random & Yes & $0.5$ \\
Active & No  & $71.2$  \\
Active & Yes & $ \mathbf{80.3}$  \\  \bottomrule
\end{tabular}
\caption{Results of the ablation study using the CUB-200-2011 dataset.\label{ablation}}
\end{table}   

\textbf{The relevance of the representative vectors extracted via the proposed model}: For this purpose, $5$ classification layers were added to the network and were fed with a different set of representative vectors as input. Precisely, instead of receiving the $3$ representative vectors, \textsl{i.e.,} $\{\hat{f}_1,\hat{f}_2,\hat{f}_3\}$, the $5$ layers only receive the following vector or combination of vectors: $\{\hat{f}_1,\hat{f}_2\}$, $\{\hat{f}_2,\hat{f}_3\}$, $\{\hat{f}_1\}$,$\{\hat{f}_2\}$ and $\{\hat{f}_3\}$. Each additional layer was trained by adding a cross-entropy term to the loss function. The sum is then divided by the total number of terms, which is 6. By doing so, the final new loss function is the average of the 6 cross-entropy terms. To prevent the supplementary layers from adjusting the feature vectors for their needs, we prevent their gradients from back-propagating to the features.   
Similar to the previous ablation study, the model with additional layers was trained in high-resolution without distillation or hyper-parameter optimization. The accuracy of each of the supplementary layers (and the main layer) is summarized in Table~\ref{tab:vecInd} on the three datasets.
As shown, when using only one vector, performance decreases along with the ranking of the feature vectors (from 1 to 3). Similarly, when removing one of the vectors from the full set, the accuracy also drops.   
This observation proves indirectly that the contribution of each representative vector to the task is related to the ranking of the feature.
Note that the performance of the model on CUB-200-2011 using the active selection method with refining from the previous ablation study ($80.3\%$) is different from the performance of the model using all three vectors $\{\hat{f}_1,\hat{f}_2,\hat{f}_3\}$. This is because, in the latter, the loss is an average of 6 terms, which means that the cross-entropy term of the model is multiplied by a $1/6$ factor, which is equivalent to using a learning rate smaller by a factor of $1/6$ compared to the previous experiment.

The impact of the choice of N on the training time is studied in the next section.

\setlength{\tabcolsep}{2pt}
\begin{table}[!htbp] 
    \centering
    \begin{tabular}{c|c|cc|ccc}
    \toprule
        Dataset & $\{\hat{f}_1,\hat{f}_2,\hat{f}_3\}$ & $\{\hat{f}_1,\hat{f}_2\}$ &  $\{\hat{f}_2,\hat{f}_3\}$ & $\{\hat{f}_1\}$ & $\{\hat{f}_2\}$ & $\{\hat{f}_3\}$  \\ \hline
        %Acc. & \textbf{79.5} & 79.1 & 77.9 & 77.3 & 73.2 & 56.8 \\
        CUB-200-2011 & \textbf{82.8}  & 82.7 & 81.4 & 81.2 & 80.0 & 66.3 \\
        Stanford cars & \textbf{88.1}& 87.3& 87.3& 85.4& 85.4& 84.0\\ 
        FGVC-Aircraft & \textbf{86.2}& 85.5& 85.4& 84.8& 83.9& 84.2\\
    \bottomrule
    \end{tabular}
    \caption{Impact of number and rank of features on accuracy.
    \label{tab:vecInd}  }
\end{table}  %   

\subsection{Training time}

To evaluate the benefit/cost ratio of using 3 vectors instead of 2 or 1, we computed the average training epoch time over 20 epochs with the default BR-NPA meta-parameter $N=3$, but also with $N=2$ and $N=1$. 
Models were also trained in high-resolution without distillation or hyper-parameter optimization.
We respectively obtained an average training epoch duration of $498.88$, $496.03$, and $480.60$ seconds for $N=3$, 2, and 1, with standard deviations of 0.65, 1.12, and 1.06.
This shows that the increase in computation time when using 3 vectors instead of 2 or 1 is limited as most of the computation time is spent on computing feature maps using the backbone ResNet-50 CNN.
Therefore using 3 vectors instead of 2 or 1 helps extract more meaningful information from the same feature maps with minimal computational overhead.

\begin{figure}
    \centering
    \includegraphics[width=0.5\textwidth]{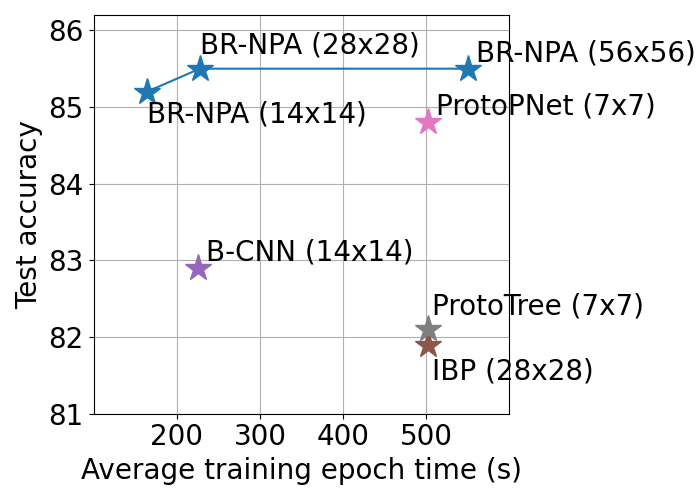}
    \caption{Test accuracy as a function of average training epoch time.
    Each point is annotated with the model name and the corresponding attention map resolution.}
    \label{trTime}
\end{figure}

We further study the training time of BR-NPA by comparing it to the training time of other models.
For simplicity, we use the same models as for fine-grained visualization, i.e. ProtoPNet, ProtoTree, and IBP.
We also measure the training time of a low-resolution BR-NPA ($14\times14$) trained without distillation and a medium resolution BR-NPA ($28\times28$) with distillation where only the stride of layer 4 is set to 1 and the stride of layer 3 is left at its default value of 2.
As in the preceding section, we computed the average training epoch time over 20 epochs.
We visualize the results in \cref{trTime} as a function of test accuracy.
This plot shows that BR-NPA is more efficient in terms of accuracy/training time ratio than the other models. 

\subsection{Increasing the resolution of feature maps} 

The effects of using a larger feature map resolution on both the performances of B-CNN \cite{B-CNN} and a BR-NPA were explored. They were both trained on the CUB-200-2011 dataset. We compared two resolutions including $14\times14$ and $56\times56$. The first is the default output resolution and the second one was obtained by reducing the strides of layers $3$ and $4$ to $1$. We used hyper-parameter optimisation and to train higher-resolution models, we employ{ed} distillation, as detailed in \cref{reducedStrideSec}.

Ideally, the attention map is expected to be sparse, \textsl{i.e.,} focusing on a small local area of the image.
To quantify this, we introduce a new metric, namely, \textit{sparsity}. First, we averaged the multiple attention maps produced by the model. Then, to take the feature vector norm into account, each element of the map was multiplied by the corresponding feature vector euclidean norm. Finally, the sparsity is defined as $s = a_{max}/a_{mean}$, where $a_{max}$ is the maximum of the resulting map and $a_{mean}$ is its mean value. An attention map focusing on a small part of the image is supposed to have a low mean value which ends out with a high sparsity value.
Conversely, a map covering all the input uniformly is supposed to have a maximum value that is close to its mean value. Therefore its sparsity is close to the minimum value, \textsl{i.e.,} $1$. 
We are therefore looking for models with a high sparsity value. 
Note that this metric is independent of the used resolution and the number of produced attention maps.
Distilling the higher-resolution model obtained via reducing strides helps to improve the sparsity of the student network compared to the teacher network ($11.2 \rightarrow 19.6$ for B-CNN, and $7.46 \rightarrow 21.3$ for BR-NPA). This shows that distillation combined with a reduction of strides helps to generate more precise, interpretable attention maps and improve the accuracy of the model. Results depicted in Table~\ref{reducedStride} show that the higher-resolution models outperform their corresponding teacher network ($+1.7$ and $+0.3$ respectively for B-CNN and BR-NPA).
For comparison, a CNN with default resolution ($14\times14$) obtains an accuracy of $84.2 \%$.  

\begin{table}[tbh!p]  
\centering
\begin{tabular}{c|c|c|c}\toprule
Model & Map size & Accuracy & Sparsity \\
\hline
%\multirow{3}{*}{CNN} &$14\times14$ & - & $84.2$&$9.52$ \\
%& $56\times56$ & \xmark & $79.2$  & $21.27$  \\
%& $56\times56$ & \cmark & $83.1$  & $\mathbf{38.4}$ \\
%\hline
%\multirow{2}{*}{B-CNN} & $14\times14$&- & $82.9$ & $11.2$  \\
\multirow{2}{*}{B-CNN} & $14\times14$ & $82.9$ & $11.2$  \\
%& $56\times56$& \xmark & $82.4$ & $37.0$   \\
%& $56\times56$& \cmark & $84.6$ & $19.6$ \\
& $56\times56$ & $84.6$ & $19.6$ \\
\hline 
%\multirow{2}{*}{BR-NPA} & $14\times14$&- & $85.2$ & $7.46$  \\
\multirow{2}{*}{BR-NPA} & $14\times14$ & $85.2$ & $7.46$  \\
%& $56\times56$ &\xmark & $82.4$ &  $4.60$ \\
%& $56\times56$ & \cmark  & $\mathbf{85.5}$  & $\mathbf{21.3}$ \\ \hline
& $56\times56$  & $\mathbf{85.5}$  & $\mathbf{21.3}$ \\ \hline
\end{tabular}
\caption{Impact of resolution on performance using the CUB-200-2011 dataset}.
\label{reducedStride} 
\end{table} 
 
\begin{figure}[!hptb]
    \centering
    \includegraphics[width=0.8\textwidth]{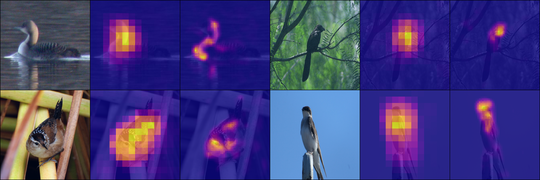}
    \caption{From left to right: original image, attention maps of CNN with lower and then higher-resolution feature maps.}
    \label{strides}
\end{figure}

Additionally, we also evaluated the impact of increasing the resolution of the feature maps on the interpretability of the attention map qualitatively. We trained a student CNN to imitate a lower-resolution teacher CNN with distillation and visualize the activation maps for both the teacher and student CNN in Fig.~\ref{strides}. It could be observed that the attention maps from the higher-resolution CNN are more interpretable as they are sparser and the activated areas are located on the sub-parts of the object instead of the whole object.
  
\subsection{Impact of the number of attention maps on accuracy}

\begin{figure}[!hptb]
    \centering
    \includegraphics[width=0.5\textwidth]{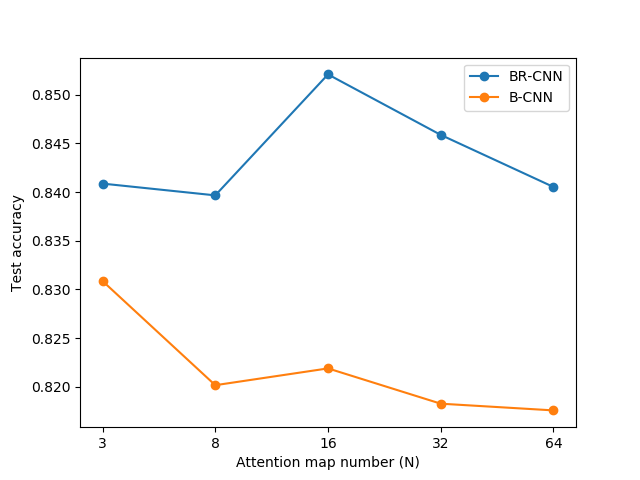}
    \caption{Impact of the attention map number $N$ on the test accuracy.}
    \label{N}
\end{figure}

The selection of the number of attention maps $N$ may influence performances. To verify this, we further examined different $N$ values and checked the corresponding accuracy of both B-CNN and BR-NPA for the task of fine-grained classification. Results are shown in Fig. \ref{N}. It is observed that by varying $N$ the performances do not change significantly for both models.
It should be noted that using values of $N$ equal to 16 or 32 leads to small accuracy gains compared to $N=3$. 
However, this multiplies the number of maps to visualize by more than 5, harming the interpretability of the model, without much accuracy improvement. As the focus of this work is interpretability we chose to use only $3$ attention maps.

Furthermore, Fig. \ref{N} shows that the value of N that maximizes B-CNN's performance is 3, which shows that using $N=3$ is not to the disadvantage of B-CNN.
% we show the test accuracy of several low-resolution B-CNN and BR-NPA trained with various values of $N$. 

\subsection{Part discriminability hierarchy \label{hierarch}}  
\begin{figure}[!hptb]
    \centering
    \includegraphics[width=0.8\textwidth]{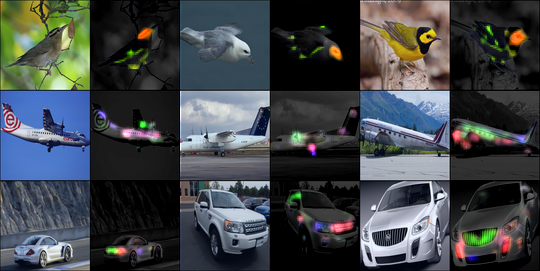}
    \caption{
    Attention maps produced by BR-NPA. 
    \label{bilVSClus}}
\end{figure}

Fig~\ref{bilVSClus} shows visually whether there is a hierarchy \textsl{i.e.,} ranking of importance level among the parts' relevance that contributes to the decision-making of the task.
For example, with the bird dataset, the first map (in red) focuses more on the head and the second map (green) focuses more on the body, indicating that the head is more discriminative than the body in the identification of the bird species. 
 
Furthermore, the blue map is not visible on the bird dataset because it has focused on vectors with  very low norms compared to ones selected by the red or green map. Low euclidean norm feature vectors have an ignorable impact on the final prediction of the task. In other words, these low-norm features are not relevant to the decision. The visualization on the aircraft dataset shows that the three maps are often visible and each can focus on various parts like tails, wheels, engines, \textsl{etc}. 
This indicates that, according to the model, it is necessary to focus on three parts as they are of equivalent importance level.
Similarly, on the cars dataset, the first and second maps focus on the front air inlet and the lights, whereas the blue maps tend to target other details like the flasher (column 2) or the car body details (column 1). It is shown that the two most important parts are the front air inlet and the lights and that some details like the body details are also less discriminative. It is showcased that BR-NPA facilitates the discovery of the importance-level hierarchy among the object's parts according to their relevance. 

\subsection{Attention metrics}

To evaluate the reliability of the proposed attention mechanism, we employed the Deletion Area Under Curve (DAUC) and Integration Area Under Curve (IAUC) metrics proposed by Petsiuk et al. \cite{rise}. 
These metrics evaluate the reliability of the saliency maps by quantifying the contribution of an area of the image in the predicted class score regarding its importance according to the saliency map.
More precisely, the DAUC metric progressively masks the whole input image and lets the model do an inference after each mask application. At each step, the variation of the initially predicted class score is measured. The intuition behind this is that if a saliency map highlights the areas that are relevant to the decision-making, masking them will result in decreasing the correct class score.
Instead of progressively masking the image, the IAUC metric starts from a blurred image and then progressively unblurs it by starting from the most important areas according to the saliency map.
Similarly, if the area highlighted by the map is relevant for predicting the correct category, the score of the corresponding class (obtained using the unblurred image) is supposed to increase rapidly.

Mean values of DAUC and IAUC over $100$ images are shown by \cref{attMetr}. It could be observed that BR-NPA yields better performance than all the other methods, including the SOTA interpretability models (\textsl{e.g.}, the IBP and ProtoTree) and the SOTA visualization methods (\textsl{e.g.}, VarGrad and SmoothGrad). 

\begin{figure}
    \centering
        \begin{subfigure}{0.47\textwidth}
        \includegraphics[width=\textwidth]{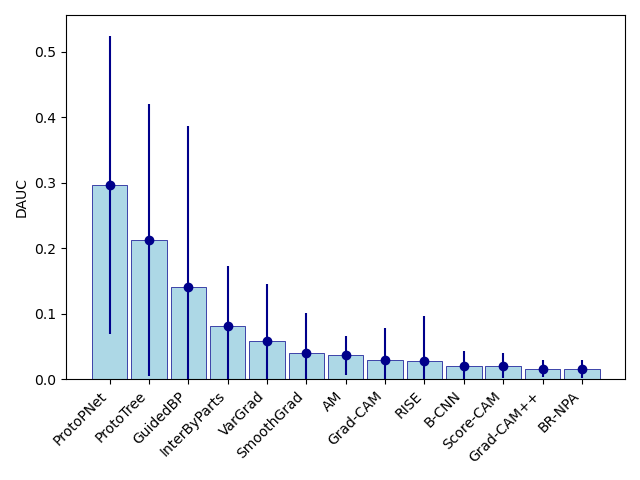}
    \end{subfigure}
    \begin{subfigure}{0.47\textwidth}
        \includegraphics[width=\textwidth]{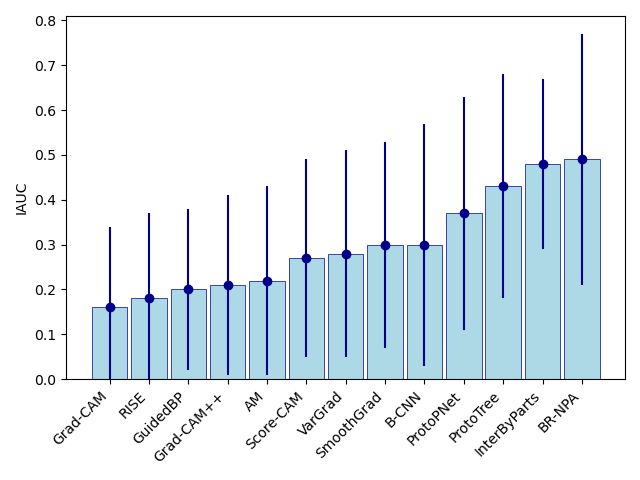}
    \end{subfigure}
    \begin{subfigure}{0.47\textwidth}
        \includegraphics[width=\textwidth]{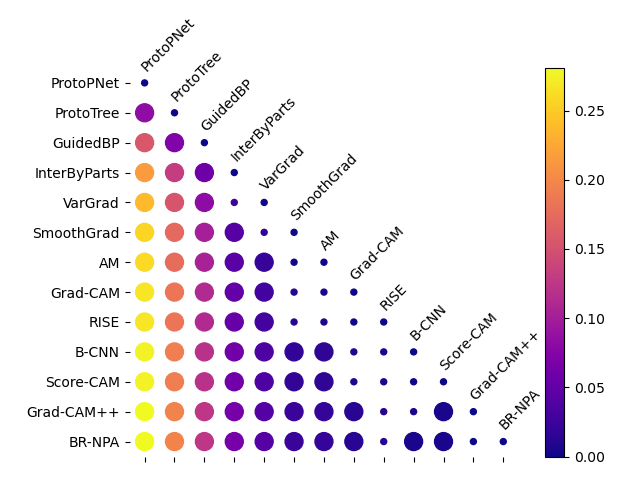}
        \caption{DAUC}
    \end{subfigure}
    \begin{subfigure}{0.47\textwidth}
        \includegraphics[width=\textwidth]{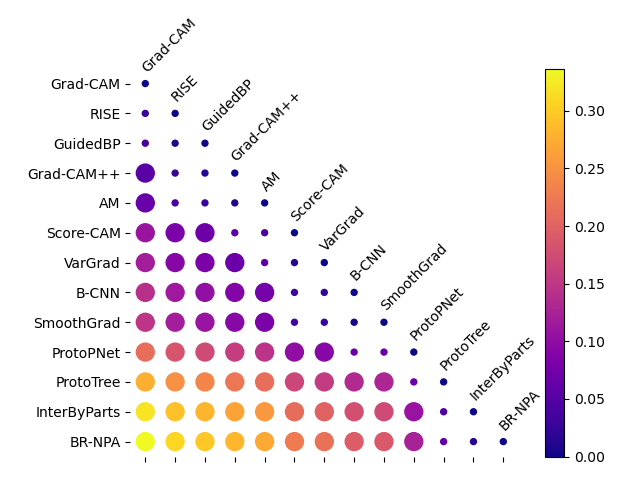}
        \caption{IAUC}
    \end{subfigure}
    \caption{
    Evaluation of the methods' attention reliability.
    Top row: Mean DAUC and IAUC with standard deviation over $100$ random test images.
    For DAUC, lower is better; for IAUC, higher is better.
    Bottom row: Mutual comparison of performance gain. Color indicates the difference in performance and size indicates statistical significance (with a $p_{val} < 0.05$ criteria), where a large disk indicates statistical significance and a small disk indicates no significance. 
    Models are listed from worst to best from top to bottom. At row $i$ column $j$ is shown the gain of method $i$ relative to $j$. }
    \label{attMetrTtest}
\end{figure}

To further verify whether the performance gains are statistically significant, we computed the Welch’s t-test between the performances of each pair of methods, and the results are visualized in \cref{attMetrTtest}. BR-NPA yields significantly better performance than most of the other methods.
The first exception is RISE in terms of DAUC which can be explained by the large standard deviation it exhibits on this metric, as shown in the top left pane of \cref{attMetrTtest}, probably preventing statistical significance, even if the score of BR-NPA is almost twice smaller than RISE's score.
The second exception is ProtoTree and IBP in IAUC which can be explained by the large variance exhibited by all models on this metric and the fact that ProtoTree and IBP obtain relatively good performance compared to BR-NPA.
However, RISE almost produce the worst performance on the IAUC metric and ProtoTree and IBP also perform poorly on the DAUC metrics. 
In general, all methods obtain different rankings according to a certain metric (DAUC or IAUC), except BR-NPA, which yields the best performance in both cases, highlighting the robustness of the attention maps explainability across several metrics.

\begin{table}[!hptb]
\centering
\begin{tabular}{c|c|c|c}\toprule
&  & DAUC  & IAUC \\ \midrule
& GuidedBP & $ 0.1406$ &$0.20$ \\ 
& VarGrad & $ 0.0578$ &$0.28$ \\ 
& SmoothGrad & $ 0.0394$ &$0.30$ \\ 
Post-Hoc& AM & $ 0.0362$ &$0.22$ \\ 
Methods& Grad-CAM & $ 0.0286$ &$0.16$ \\ 
& RISE & $ 0.0279$ &$0.18$ \\ 
& Score-CAM & $ 0.0207$ &$0.27$ \\ 
& Grad-CAM++ & $ 0.0161$ &$0.21$ \\ 
\hline
 & ProtoPNet & $ 0.2964$ &$0.37$ \\ 
\multirow{2}{*}{Attention}& ProtoTree & $ 0.2122$ &$0.43$ \\ 
\multirow{2}{*}{Architectures}& B-CNN & $ 0.0208$ &$0.30$ \\ 
& IBP & $ 0.0811$ &$0.48$ \\ 
\cline{2-4}
& BR-NPA & $ \mathbf{0.0155 }$ &$\mathbf{0.49}$ \\ 
\end{tabular}
\caption{Evaluation of the saliency methods' reliability. For DAUC, lower is better; for AUC, higher is better. Each value is the mean over the same $100$ randomly selected test images.\label{attMetr}}
\end{table}

%\subsection{Memory footprint and efficiency} 
%Last but not least, BR-NPA has a smaller memory footprint and requires less inference time compared to B-CNN for various backbones, while obtaining better accuracy. Relevant results and analysis are  in the supplemental materials.

\section{Conclusion}
In this work, we proposed a non-parametric attention module that can be easily integrated into an existing architecture to significantly improve its interpretability while maintaining its accuracy.
Specifically, this method improves the usual neural attention mechanism in two ways.
First, the feature maps resolution was increased using distillation to highlight finer details. Second, the neural layer was replaced by a well-designed non-parametric module such that the importance level of the parts composing the object can be visualized using color. 
Through quantitative and qualitative experiments, it is demonstrated that BR-NPA achieves more comprehensive and accurate visual explanations compared to SOTA attention models and visualizations methods across multiple tasks without compromising the classification accuracy.
Furthermore, it is an easy plugin that could be applied to various problems like fine-grained classification, re-identification, and few-shot classification and produces saliency maps matching the task objective. 
Contrarily to the usual attention, it is easy to integrate it into a complex pretrained architecture without significant accuracy loss.

We have mentioned that the exact implication of concatenating the representative vectors is yet to be explored.
An alternative to concatenation could be to use one linear model per representative vector before aggregating the models' predictions.
This would allow using large numbers of representative vectors per image without risking any optimization issues.
However, this would limit the ability to model interactions between features from different parts and we showed in \cref{tab:vecInd} that these interactions help to improve performance.
An improved version of BR-NPA would allow modeling these interactions while maintaining efficient optimization.

We focused our evaluation on objective metrics (Accuracy, DAUC/IAUC) and qualitative evaluation.
Therefore, in the future, we will run a subjective study to investigate the use of BR-NPA compared to other frameworks in understanding medical computer-assisted decisions.
Indeed, we can expect for example that higher-resolution maps help remove ambiguities about the actual focus of the model, which probably helps the user to construct a more accurate mental model of the model. 
However, a user study is still necessary to understand the impact of BR-NPA on users compared to other frameworks and in particular, measure how much it affects the acceptability of the automatic decision and the user's trust in the model.

\section{Acknowledgments}
This study was partly supported by NExT (ANR-16-IDEX-0007). The authors have no conflict of interests to declare.

\bibliographystyle{elsarticle-num}
\bibliography{biblioShort}

\newpage
\appendix

\section{Introduction}
This supplementary material provides more contents including: (1) an extended characterization of BR-NPA with a study of the impact of the proposed feature selection and refinement strategies on visualizations; 
%(2) results that showcase the advantages of the proposed BR-NPA in terms of efficiency and memory footprint; 
%(3) extensive experimental results of BR-NPA on semantic segmentation, which demonstrate again the feasibility of plugging the proposed framework into complex pretrained network architecture; 
%(4) more experimental results compared to state-of-the-art attention model designed for person re-identification; 
(2) extra visualization results for the considered tasks mentioned in the paper. 
%along with novel visualized results for task of semantic segmentation.}

%Finally, extended visualizations showing the flexibility of BR-NPA are proposed on all the tasks mentioned in this paper, including semantic segmentation. 
% we compare the performance and attention maps of BR-NPA with the Relation-aware Global Attention (RGA) proposed by \cite{relationAwareReID}, an attention model proposed for person re-identification.  we compare the performance and attention maps of BR-NPA with the Relation-aware Global Attention (RGA) proposed by \cite{relationAwareReID}, an attention model proposed for person re-identification. 

\section{Hyper-parameters values for ablation study}

%\paragraph{Fine-grained classification.} For all fine-grained experiments on fine-grained classification we used a ResNet-50 \cite{resnet} trained during 50 epochs with $448\times448$ images. 
%During training, images are transformed with random crop, horizontal flip and color jitter and then normalized according to the usual ImageNet procedure \cite{imagenet}. 
For the ablation study, we used a learning rate of 0.001, an SGD optimizer with a momentum of 0.9, no weight decay, no scheduler, no dropout, a batch size of 12, and a crop ratio of 0.875.
%For the other experiments, we optimised learning rate, optimizer choice (among SGD, Adam \cite{adam} and AdaGrad \cite{adagrad}), momentum, weight decay, dropout, batch size

\section{BR-NPA characterization}

%\section{Extended ablation study}

\subsection{Visualization of the Ablative Models}
\begin{figure}[!hptb]
    \centering
    \includegraphics[width=\textwidth]{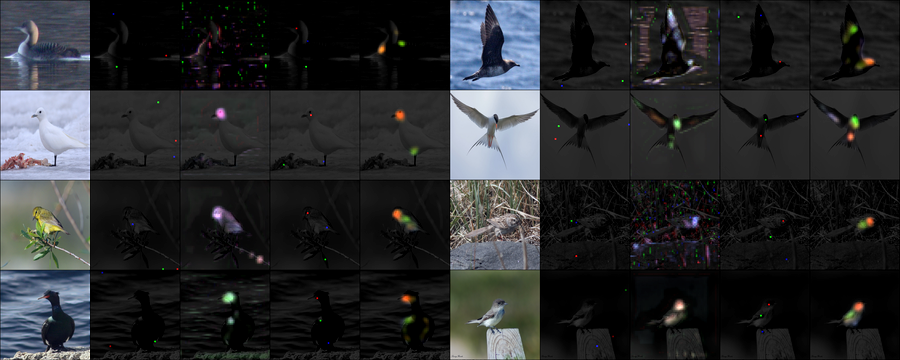}
    \caption{From left to right: the original image, a BR-NPA selecting random vectors without refinement, a BR-NPA selecting random vectors, a BR-NPA without refinement, and a BR-NPA selecting high norm vectors and refining them. Selecting random vectors leads to focus on irrelevant parts of the input and not refining them makes it difficult to understand which part the model has focused on.}
    \label{fig:ablation}
\end{figure}

In this section, more visually friendly results of the ablation studies regarding the proposed active feature selection and refinement strategy are provided. Table~\ref{ablation} recalls the performance of the $3$ variants: (1) the ablative model, where the feature vectors are uniformly and randomly selected among all available vectors; (2) the ablative model without the refinement phase; (3) the ablative model with random selection and no refinement phase. Attention maps obtained with the 3 ablative models are depicted in Fig.~\ref{fig:ablation}. As shown, selecting random vectors, instead of the high norm ones, ends out in highlighting parts that are not contributing to decision makings (\textsl{i.e.,} parts that are not imperative for classification). Moreover, without feature refinement, it is difficult to interpret which of the model is contributing.

%shows that selecting random vectors instead of high norm ones leads to focus on parts of the input that are not relevant for classification and not refining them makes it difficult to understand on which part the model has focused.

\begin{table}[!hptb]  
\small{ 
\centering
\begin{tabular}{c|c|c}\toprule
Raw vector selection method & Refining & Accuracy   \\ 
\hline
Random & No & $0.5$ \\
Random & Yes & $0.5$ \\
Active & No  & $71.2$  \\
Active & Yes & $ \mathbf{80.3}$  \\  \bottomrule
\end{tabular}
\caption{Results of ablation study.\label{ablation}}}
\end{table}

\section{More Visualisation Results}

%In Figs.~\ref{fine,reid,fewShot} are shown more visual comparison with Guided Grad-CAM++, Grad-CAM, Grad-CAM++, AM, B-CNN and BR-NPA. 

\begin{figure*}[!htbp] 
    \centering
    \includegraphics[height=\textheight]{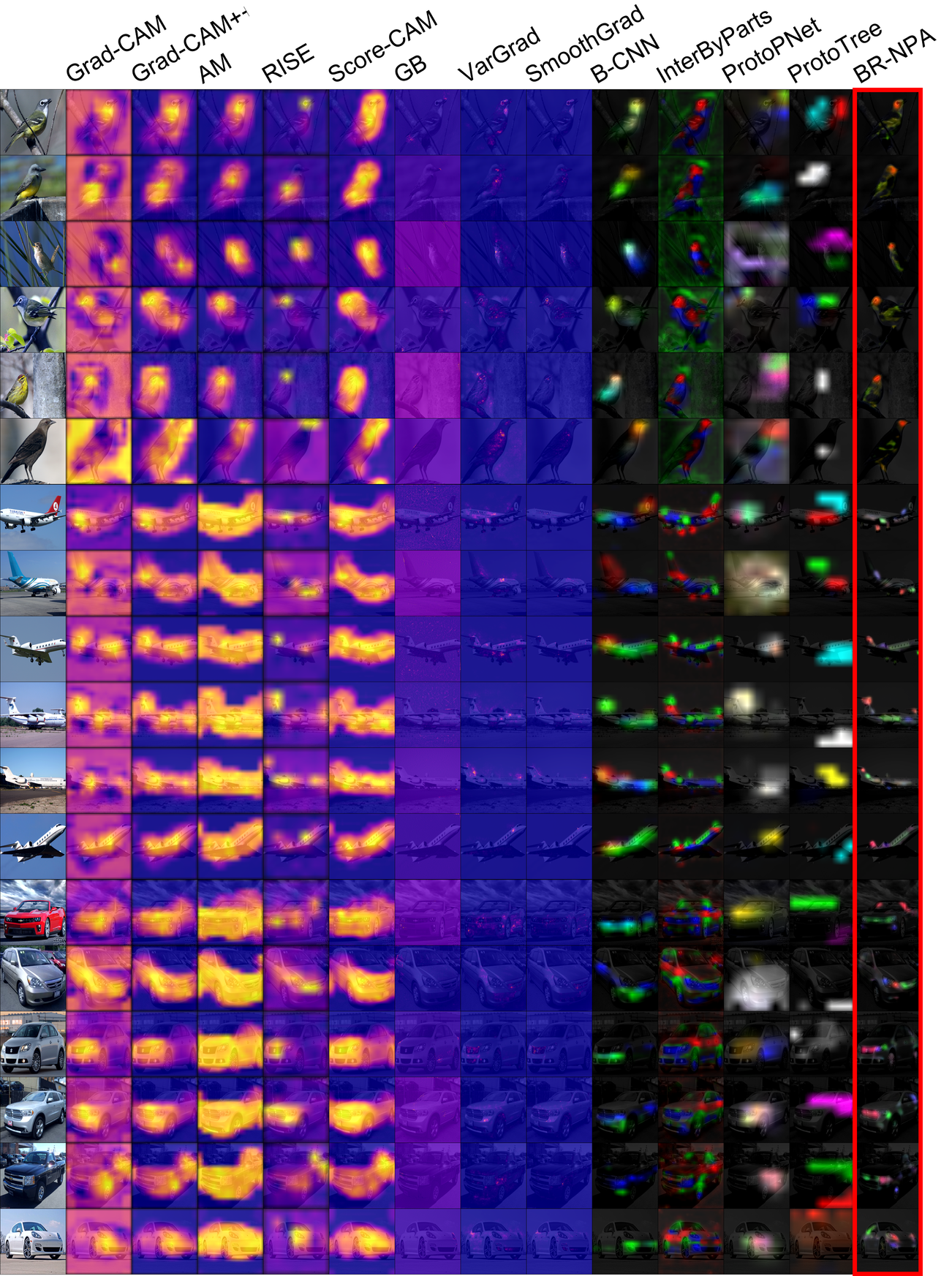}
    \caption{Comparison of different visual explanations with BR-NPA for fine-grained classification on 3 datasets: CUB-200-2011  lines 1-6, FGVC-Aircraft lines 7-12, and the Standford cars lines 13-18. BR-NPA yields detailed attention maps focused on semantic parts of the object whereas the other methods either produce blurry maps covering the whole object or extremely sparse maps making it difficult to identify the important areas\label{fine}.}
\end{figure*}

\begin{figure*}[!htbp] 
    \centering
    \includegraphics[height=\textheight]{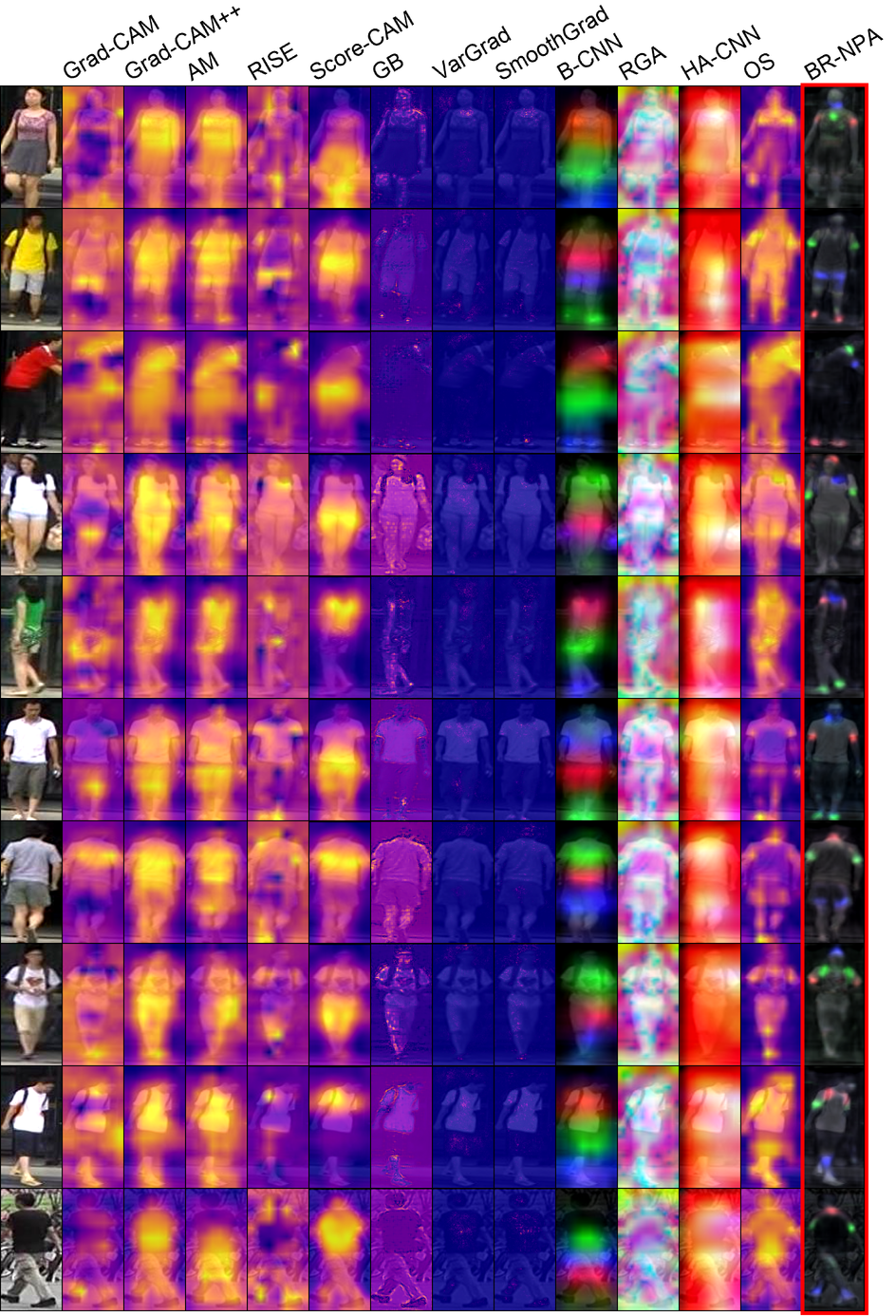}
    \caption{Comparison of different visual explanations with BR-NPA for person re-identification task on the Market-1501 dataset \label{reid}}.
    
\end{figure*}

\begin{figure*}[!htbp] 
    \centering
    \includegraphics[height=\textheight ]{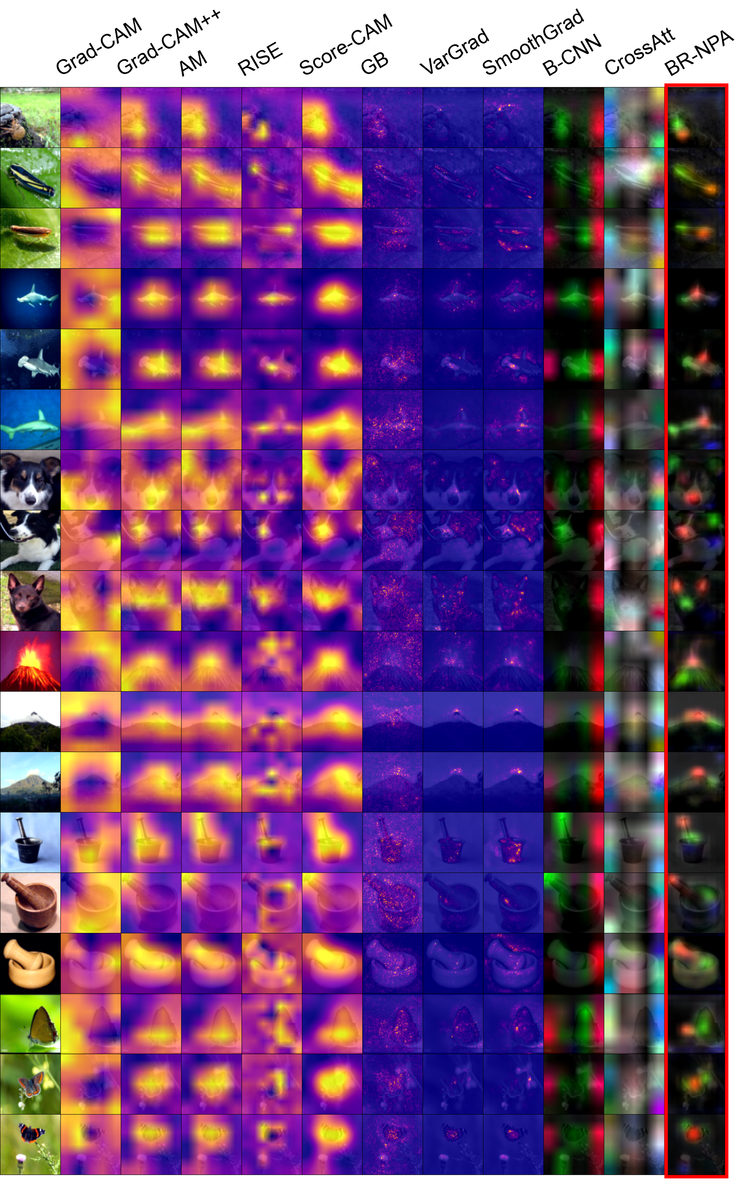}
    \caption{Comparison of different visual explanations with BR-NPA for the few-shot classification task on the TieredImagenet dataset.
    \label{fewshot}}
\end{figure*}

Supplementary visualization of the attention maps are presented in Figs.~\ref{fine},\ref{reid} and \ref{fewshot}. %Among all the images, the attention maps produced by BR-NPA highlight more precise details of the objects that capture well the inter-class unique characteristic. For example, in rows 4-6, highlighted finer-grained local regions are most of the time the tails or wings of the airplanes, which help to differ different airplane models. Similarly, B-CNN is also able to identify similar local important parts, but they are generally less precise. Conversely, the highlighted regions from Grad-CAM++ and AM approaches are significantly coarser, which cover the entire objects, while the ones from Guided Grad-CAM++ and Grad-CAM are sparse and spread out randomly all over the images. None of those attention regions provide precise information that helps the decision-making of fine-grained classification.

%Obviously, the attention maps obtained from different tasks highlight the target object at different granularity-level. For the task of fine-grained classification and person re-identification, our method yields more precise attention maps that focus on finer object parts. Differently, for few-shot classification, the proposed model tends to focus on the entire object as the objective of few-shot is to distinguish between categories instead of sub-categories with only a few samples per unseen class. It is demonstrated that our model is able to generate interpretable visualized maps according to the goal of the task, which could provide important insights for the development of different methodologies, \textsl{i.e.,} different network architectures that pay more attention to a certain type of local areas, regarding the goal of the target task.

\end{document}